\documentclass[sigconf]{acmart}
\AtBeginDocument{%
  \providecommand\BibTeX{{%
    \normalfont B\kern-0.5em{\scshape i\kern-0.25em b}\kern-0.8em\TeX}}}

\usepackage{graphicx}

\usepackage{booktabs}
\usepackage{multirow}
\usepackage{enumitem}
\usepackage{physics,amsmath}
 
\usepackage{amssymb}
\usepackage{colortbl}
\newcommand{\wx}[1]{{\color{black}{#1}}}

\newcommand{\lyc}[1]{{\color{black}{#1}}}
\newcommand{\imp}[1]{{\color{red}{#1}}}
\newcommand{\white}[1]{{\color{white}{#1}}}
\definecolor{gray}{rgb}{ 0.851,  0.851,  0.851}

\newcommand{\ie}{\emph{i.e., }}
\newcommand{\eg}{\emph{e.g., }}

\newcommand{\st}{\emph{s.t. }}

\newcommand{\cf}{\emph{cf. }}

\newcommand*\rot{\rotatebox{90}}

\copyrightyear{2022}
\acmYear{2022}
\setcopyright{rightsretained}
\acmConference[MM '22]{Proceedings of the 30th ACM International Conference on Multimedia}{October 10--14, 2022}{Lisboa, Portugal}
\acmBooktitle{Proceedings of the 30th ACM International Conference on Multimedia (MM '22), Oct. 10--14, 2022, Lisboa, Portugal}
\acmDOI{10.1145/3503161.3548035}
\acmISBN{978-1-4503-9203-7/22/10}
\acmPrice{15.00}

\usepackage{etoolbox}
\makeatletter
\patchcmd{\maketitle}{\@copyrightpermission}{
   \begin{minipage}{0.3\columnwidth}
     \href{https://creativecommons.org/licenses/by/4.0/}{\includegraphics[width=0.90\textwidth]{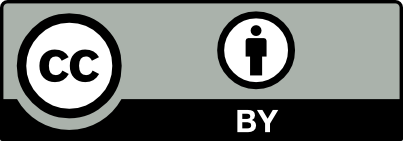}}
   \end{minipage}\hfill
   \begin{minipage}{0.7\columnwidth}
     \href{https://creativecommons.org/licenses/by/4.0/}{This work is licensed under a Creative Commons Attribution International 4.0 License.}
   \end{minipage}

   \vspace{5pt}
}{}{}

\makeatother

\author{Yicong Li$^{1}$, Xiang Wang$^{2*}$, Junbin Xiao$^{1}$, and Tat-Seng Chua$^1$}
\def\authors{Yicong Li, Xiang Wang, Junbin Xiao, and Tat-Seng Chua}
\affiliation{
\institution{$^1$National University of Singapore, $^2$University of Science and Technology of China}
\country{}
}
\email{liyicong@u.nus.edu, xiangwang1223@gmail.com, junbin@comp.nus.edu.sg, dcscts@nus.edu.sg}
\thanks{$*$ Corresponding author. This research is supported by the Sea-NExT Joint Lab, and the CCCD Key Lab of Ministry of Culture and Tourism, USTC.}
     
\renewcommand{\shortauthors}{Li et al.}

\begin{document}

\title{Equivariant and Invariant Grounding for Video Question Answering}

\begin{abstract}
Video Question Answering (VideoQA) is the task of answering the natural language questions about a video.
Producing an answer requires understanding the interplay across visual scenes in video and linguistic semantics in question.
However, most leading VideoQA models work as black boxes, which make the visual-linguistic alignment behind the answering process obscure.
Such black-box nature calls for visual explainability that reveals ``What part of the video should the model look at to answer the question?''.
Only a few works present the visual explanations in a post-hoc fashion, which emulates the target model's answering process via an additional method.
Nonetheless, the emulation struggles to faithfully exhibit the visual-linguistic alignment during answering.

Instead of post-hoc explainability, we focus on intrinsic interpretability to make the answering process transparent.
At its core is grounding the question-critical cues as the causal scene to yield answers, while rolling out the question-irrelevant information as the environment scene.
Taking a causal look at VideoQA,
we devise a self-interpretable framework, \underline{E}quivariant and \underline{I}nvariant \underline{G}rounding for Interpretable \underline{V}ideoQA (EIGV).
Specifically, the equivariant grounding encourages the answering to be sensitive to the semantic changes in the causal scene and question; in contrast, the invariant grounding enforces the answering to be insensitive to the changes in the environment scene.
By imposing them on the answering process, EIGV is able to distinguish the causal scene from the environment information, and explicitly present the visual-linguistic alignment.
Extensive experiments on three benchmark datasets justify the superiority of EIGV in terms of accuracy and visual interpretability over the leading baselines.
Our code is available at \url{https://github.com/yl3800/EIGV}.

\end{abstract}

\ccsdesc[500]{Information systems~Question answering}
\ccsdesc[500]{Information systems~Multimedia and multimodal retrieval}

\keywords{\lyc{Video Question Answering, Invariant Learning, Equivariant Learning, Interpretability}}

\maketitle
\begin{sloppypar}

\section{Introduction}
\label{sec:introduction}

\begin{figure}[t]
\centering
\includegraphics[scale=0.47]{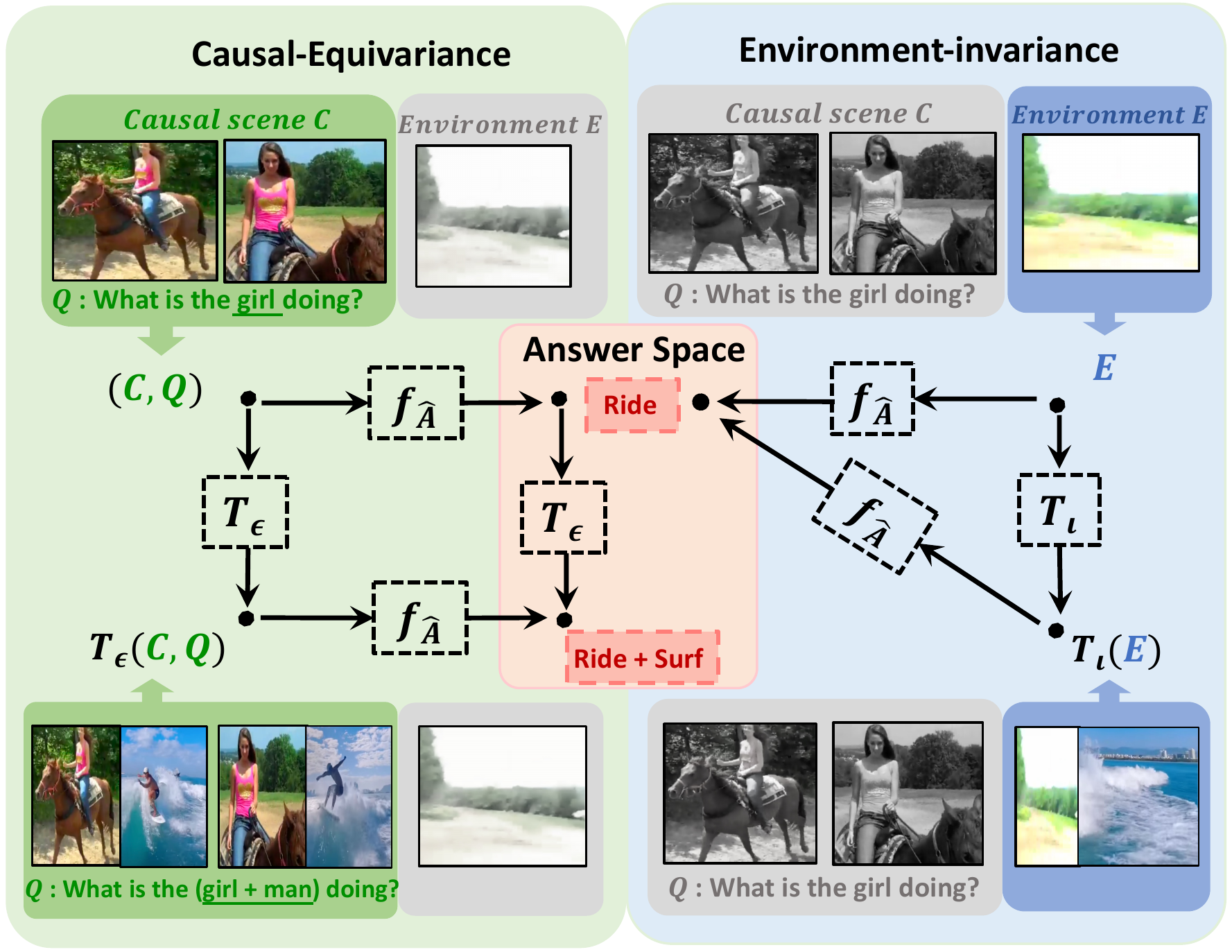}
\vspace{-15pt}
\caption{
Illustration of equivariant and invariant grounding.
The causal-equivariant principle (left) asks that the semantic change $T_{\epsilon}$ applied to the causal scene $C$ and question $Q$ should be faithfully reflected in the answer change.
In contrast, the environment-invariant principle (right) outputs the same answer, regardless of changes $T_{\iota}$ on the environment scene $E$.
Here, $f_{\hat{A}}$ maps input to answer space.
}
\vspace{-15pt}
\label{fig:overview}
\end{figure}

Video Question Answering (VideoQA) \cite{zhong2022video} is a keystone in interactive AI, such as vision-language navigation and communication systems.
It aims to answer the natural language question based on the video content.
Striving for the architecture novelty, many studies have been conducted on modeling VideoQA's multi-modal nature, such as fostering the vision-language alignment \cite{jiang2020reasoning,park2021bridge} and revisiting the visual input structure \cite{le2021hierarchical,dang2021hierarchical}.
However, existing VideoQA models usually operate as black boxes, which fail to exhibit the working mechanism behind the predictions and hardly exhibit ``What knowledge should the model use to answer the question about the video?''.
As a result, the black-box nature causes concern for the model's reliability, especially in applications to safety and security.

The concern on the black-box nature calls for  \lyc{better transparency} of VideoQA models.
Here we focus on visual-explainability \cite{CSS,DBLP:conf/ijcai/RossHD17}, aiming to reveal ``Which part of the video should the model look at to answer the question?''.
It requires us to find a subset of visual scenes --- rationale --- \lyc{that support the answering as evidence in way of human interpretation} \cite{DBLP:conf/ijcai/RossHD17}.
Taking Figure \ref{fig:overview} as an example, when answering the question ``What is the girl doing?'', the rationale should focus on the ``girl-riding on-horse'' scene in the first two clips.
Towards this end, existing studies \cite{gao2018motionappearance,DBLP:conf/iccv/Liu0WL21,DBLP:conf/mm/WangG0W21} dwell mainly on the paradigm of \textbf{post-hoc explainability} \cite{LIME,DBLP:conf/iccv/SelvarajuCDVPB17}, which distributes the predictive answer of the target model to the input visual features via an additional explainer method.
They visualize the attention weights or gradient-like signals toward the visual features, and then identify a salient pattern as the rationale.
However, post-hoc explainability has several major limitations:
(1) It fails to make the target model intrinsically interpretable \cite{DBLP:conf/cvpr/YangZQ021,wang2021causal,DBLP:journals/natmi/Rudin19},  only approximating the decision-making process of the model.
As a result, the identified rationale cannot faithfully reveal how the model leverages the multi-modal information.
(2) Such visual inspections are fragile against input perturbations, since some artifacts can be easily captured as explanations instead of genuine knowledge from the data \cite{DBLP:conf/ijcai/LaugelLMRD19,slack2020fooling,heo2019fooling,ghorbani2019interpretation}.

The limitations of post-hoc explainability inspire us to explore the paradigm of \textbf{intrinsic interpretability} \cite{ghorbani2019interpretation,DBLP:journals/natmi/Rudin19}, which embeds a rationalization module into the model to make the decision-making process transparent.
Surprisingly, the intrinsic interpretability of VideoQA models is until-now lacking.
To fill the void, we draw on \textbf{causal theory} \cite{pearl2016causal,pearl2009causal} to formulate the interpretability task as disclosing ``Which part of the video is critical/causal to answering the question?''.
Concretely, we aim to identify the causal component of input video on-the-fly, which holds the question-response information and filters out the question-irrelevant cues.
Following this essence, one straightforward realization is to ground the input video into two segments:
(1) \textbf{causal scene}, which retains the question-critical visual content and sufficiently approaches the answer, thus naturally serving as the rationale;
and (2) \textbf{environment scene}, which holds the question-irrelevant visual content and can be seen as the rationale's complement.

%
However, discovering causal scene without the supervision of ground-truth rationale is challenging.
With a causal look at the reasoning process (\cf Section \ref{sec:causal-view}), we argue that the crux of intrinsic interpretability is to amplify the connection between the causal scene and the answer, while blocking the non-causal effect of the environment scene.
Following this line, we propose two principles to guide the grounding of the rationale:
\begin{itemize}[leftmargin=*]
    \item \textbf{Causal-Equivariance.}
    By ``equivariance'', we mean that answering should be sensitive to the semantic changes on the causal scene and question (termed E-intervention), \eg any change on the causal scene and question should be faithfully reflected on the predicted answer. For example, in Figure \ref{fig:overview}, the ``girl-riding on-horse'' and ``man-surfing in-ocean'' scenes are the oracle rationales of ``What is the girl doing?'' and ``What is the man doing?'', respectively. The intervention \cite{li2021interventional} applied on the input (\ie mixing the ``girl-riding on-horse'' and ``man-surfing in-ocean'' scene, and combining two questions as ``What is the girl doing? What is the man doing?'') should set off an equivariant change in the answer (\ie changing from ``Ride'' to ``Ride+Surf'').

    \item \textbf{Environment-Invariance.}
    By ``invariance'', we mean that answering should be insensitive to the changes in the environment scene (termed I-intervention), conditioning on the causal scene and question.
    Considering Figure \ref{fig:overview} again, the intervention applied to the environment (\ie mixing the ``meadow'' and ``ocean'' scenes) implies no impact towards answering ``What is the girl doing?'', reflecting a homogeneity in the answer space.
\end{itemize}

To impose these two principles for intrinsic interpretability, we propose a new framework, \underline{E}quivariant and \underline{I}nvariant \underline{G}rounding for Interpretable \underline{V}ideoQA (\textbf{EIGV}).
EIGV equips the VideoQA backbone model with three additional modules:
a grounding indicator, an intervener, and a disruptor.
First, the grounding indicator learns to attend the causal scene based on the input question, while leaving the rest as the environment.
Then, the intervener parameterizes the proposed principles to guide the grounding.
Specifically, towards the causal-equivariance principle, it conducts the E-intervention on the causal scene and question --- that is, mix them with the counterparts from another video-question pair --- and encourages the predictive answer to be anticipated accordingly.
Towards the environment-invariance principle, when leaving the causal scene and question untouched, it applies the I-intervention on the environment --- that is, mix it with the environmental stratification of a memory bank --- and enforces the predictive answer to be invariant.
Moreover, we build an unified sight of two principles via the lens of contrastive learning.
Concretely, on top of each intervened video-question pair, the disruptor constructs the positive views by disrupting the environment scene randomly, while creating the negative views by substituting the causal scene with random scenes.
Training with these two principles allows the backbone model to distinguish the causal scene from the environmental cues, and hinge on the critical visual-linguistic alignment.

Briefly put, our contributions are: 
\begin{itemize}[leftmargin=*]
    \item We propose EIGV, a model-agnostic VideoQA framework that distills the causal visual-linguistic alignment to generate answers in a self-interpretable manner.
    
    \item We investigate the soundness of grounding rationale by posing the equivariant-invariant principle on visual grounding.
    
    \item We justify the superiority of EIGV on three popular benchmark datasets (\ie MSVD-QA \cite{DBLP:conf/mm/XuZX0Z0Z17}, MSRVTT-QA \cite{DBLP:conf/mm/XuZX0Z0Z17},  NExT-QA \cite{DBLP:conf/cvpr/XiaoSYC21}) with extensive experiments, where our design outmatches the state-of-the-art models. Moreover, our EIGV is a model-agnostic framework that can be applied to different VideoQA models. 
\end{itemize}

\section{Preliminaries}
\label{sec:preliminaries}

Here we provide a holistic view of VideoQA by summarizing a common paradigm throughout existing works. Specifically, we denote a variable and its deterministic value by upper-cased (\eg $A$) and lower-cased (\eg $a$) letters, respectively. 

\vspace{5pt}
\noindent\textbf{Modeling}.
Given the video $V$, the VideoQA model $f_{\hat{A}}{(\cdot)}$ answers the question $Q$ by formulating the visual-linguistic alignment: 
\begin{gather}\label{eq:conventional-modeling}\
    \hat{A} = f_{\hat{A}}(V,Q),
\end{gather}
where $\hat{A}$ is the predictive answer. Typically, $f_{\hat{A}}{(\cdot)}$ is a combination of two modules: 
\begin{itemize}[leftmargin=*]
    \item Video-question encoder, which warps up the visual content and linguistic semantics via two encoders: (1) the video encoder capsules the video content by methods like hierarchical design \cite{DBLP:conf/mm/PengYBW21, dang2021hierarchical, le2021hierarchical}, enhanced memory architecture \cite{gao2018motionappearance, fan2019heterogeneous} and structural graph representation  \cite{jiang2020reasoning,huang2020locationaware,DBLP:conf/acl/GuoZJ0L20, Wang_2018_ECCV}; (2) the question encoder embeds the contextual information into linguistic representation through multi-scale semantic integration \cite{jiang2020reasoning, DBLP:conf/acl/SeoKPZ20, 2021} or grammatical dependencies parsing \cite{park2021bridge}.
    
    \item Answer decoder, which abridges the encoded visual-linguistic information via cross-modal interaction methods like graph alignment \cite{ park2021bridge} and progressive attention \cite{DBLP:conf/acl/SeoKPZ20, DBLP:conf/mm/PengYBW21}, then generates the prediction accordingly.
\end{itemize}

\vspace{5pt}
\noindent\textbf{Learning}.
\wx{To optimize the video-question encoder and answer decoder, current VideoQA models usually adopt the scheme of empirical risk minimization (ERM) \cite{gao2018motionappearance,le2021hierarchical,jiang2020reasoning,DBLP:conf/mm/PengYBW21}, which measures and minimizes the risk between the ground-truth answer $A$ and predictive answer $\hat{A}$:}
\begin{gather}\label{equ:erm-loss}
    \min\mathcal{L}_{\text{ERM}}(\hat{A}, A).
\end{gather}
In essence, ERM recklessly takes the video content as a whole and enforces the risk deduction over compassion of question and every video frame, which \wx{hardly discovers a reliable interpretation to exhibit the visual-linguistic alignment}.



\section{VideoQA Reformulation}
\label{sec:reformulation}

\wx{Here we argue that disclosing ``Which part of the video is critical to answering the question?'' is the key to presenting the visual-linguistic alignment explicitly.
To this end, we take a causal \cite{pearl2009causal} look at the reasoning process of VideoQA, then formalize it as a Structure Causal Model (SCM) \cite{pearl2016causal} by investigating the causal relationships among five variables: input video $V$, question $Q$, causal scene $C$, environment scene $E$, ground-truth answer $A$.
}


\subsection{Causal Graph of VideoQA}\label{sec:causal-view}
\begin{figure}[t]
\centering
\includegraphics[scale=0.3]{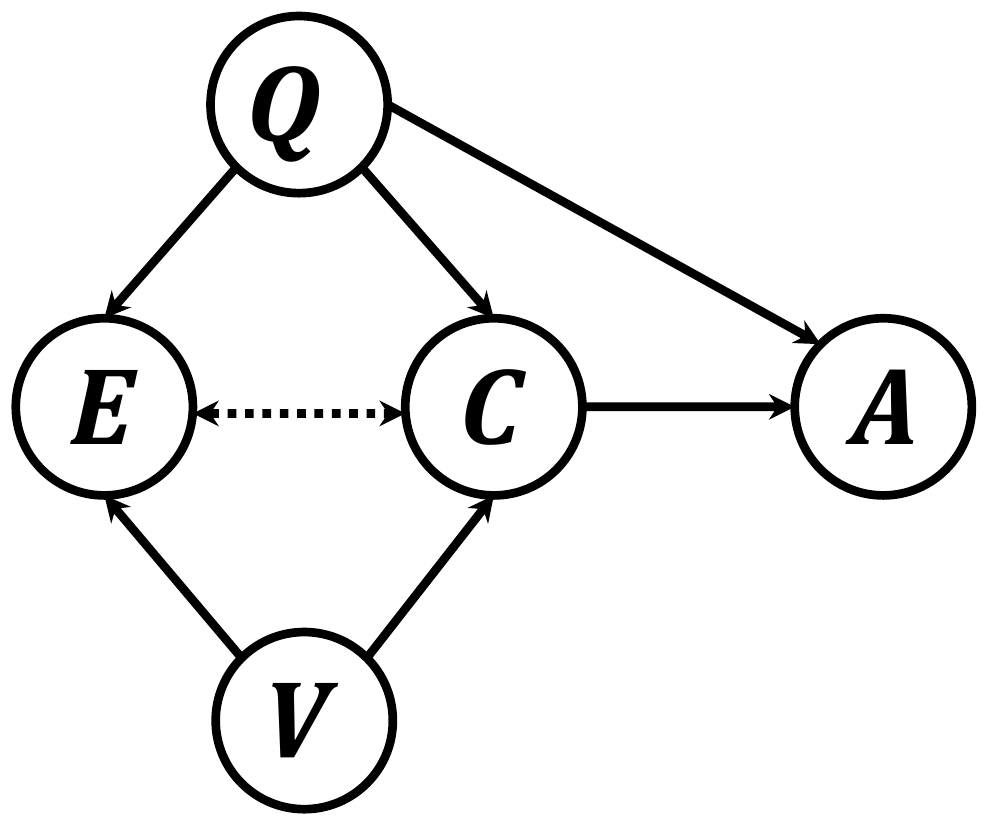}
\vspace{-5pt}
\caption{Causal Graph of VideoQA}
\vspace{-10pt}
\label{fig:scm}
\end{figure}

\wx{Figure \ref{fig:scm} illustrates the causal graph, where each link depicts the cause-effect relationship between two variables:
\begin{itemize}[leftmargin=*]
    \item $Q\to C, E \gets V$. Given the question of interest $Q$, the video $V$ can be partitioned into two parts: (1) the causal scene $C$, which retains the question-critical information and naturally serves as the rationale for answering, (2) the environment scene $E$, which gathers the cues irrelevant to the question-answering. For example, to answering ``What is the girl doing?'' in Figure \ref{fig:overview}, $C$ should be the first two clips describing the ``girl-riding on-horse'' scene, while $E$ should be the last clip about the ``meadow'' scene. Moreover, the varying semantics of different questions will emphasize different $C$.
    \item $Q\rightarrow A \leftarrow C$. The visual knowledge in the causal scene $C$ and the linguistic semantics in the question $Q$ collaborate together to determine the answer $A$. Furthermore, this path, which presents the visual-linguistic alignment, internally interprets the reasoning.
    \item $E\dashleftarrow\dashrightarrow C$. The dashed arrow sketches additional probabilistic dependencies \cite{reason:Pearl09a} between $C$ and $E$, which typically arise from selection bias \cite{DBLP:conf/cvpr/TorralbaE11}. For example, the ``meadow'' scene is frequently collected as a common environment for the ``horse-riding'' scene. 
\end{itemize}}

\subsection{Beyond ERM}
\lyc{With inspections on prior VideoQA studies, we investigate their inability to distinguish the causal and non-causal effects of scenes.
Specifically, in conventional VideoQA models, video and question are directly paired together to model their interaction and approach the golden answer, consequently.
Inevitably, taking video as a whole leaves the contributions of scenes untouched, thus failing to differentiate $C$ from $E$ and forgoing their function divergence towards the answer.
Worse still, ERM enforces these models to blindly capture the statistical correlations between the video-question pairs and answers.
As such, the visual-linguistic alignment hinges easily on the spurious correlations between $E$ and $A$, owing to the backdoor paths \cite{pearl2016causal}, which hinders the generalization of models \cite{niu2020counterfactual,wu2022dir}.
Therefore, identifying the causal scene $C$ is the critical to addressing these limitations.}

\section{Methodology}
\label{sec:method}

\begin{figure*}[t]
\centering
\includegraphics[width=0.95\textwidth]{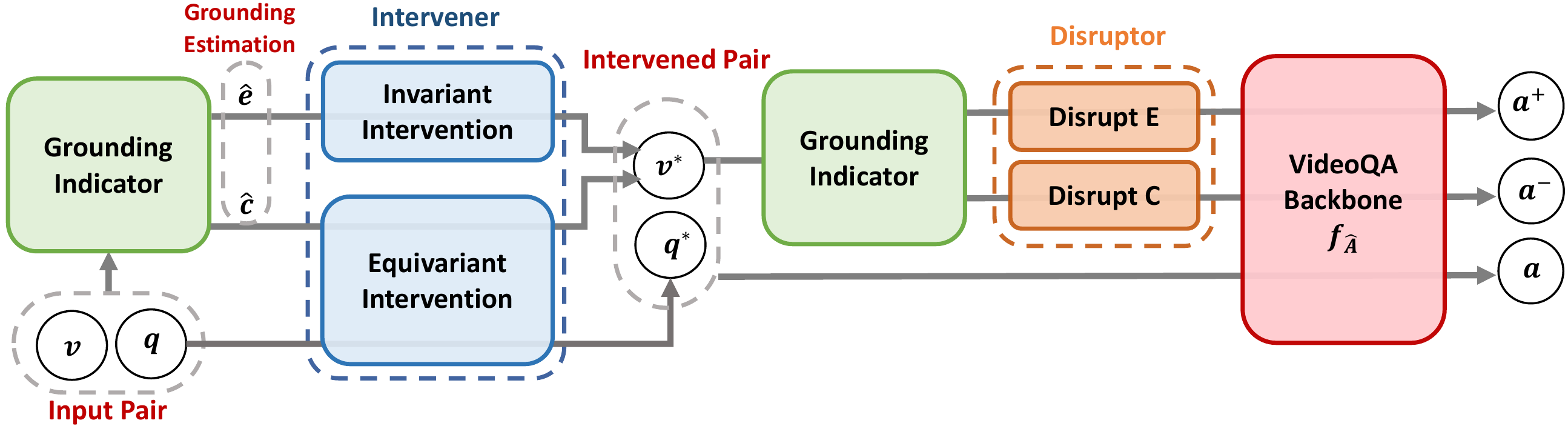}
\vspace{-14pt}
\caption{Overview of EIGV. It comprises three additional modules on top of the conventional VideoQA backbone: 1) Grounding indicator, 2) Intervener, and 3) Disrupter. First, the grounding indicator learns the estimation of causal scene $\hat{c}$ and environment $\hat{e}$. Next, two interventions are imposed on the causal and non-causal factors to compose the intervened pair $(v^*,q^*)$. Finally, based on the re-grounded result, the disruptor creates contrastive samples, which are further feed into the VideoQA backbone.}
\vspace{-5pt}
\label{fig:main}
\end{figure*}

\wx{
To ground the causal scene $C$ in the video $V$, we take a closer look at the VideoQA SCM (\ie Figure \ref{fig:scm}a), and emphasize the essential differences between $C$ and $E$.
Specifically, given the causal scene $C$ and question $Q$, the answer $A$ is determined, regardless of the variations in the environment scene $E$:}
\begin{gather}\label{equ:eq0}
      A\bot E \mid C,Q,
\end{gather}
where $\bot$ denotes the probabilistic independence. 

\vspace{5pt}
\noindent\textbf{Rationalization.} 
During training, the oracle grounding rationale $C$ is out of reach, while only the input $(V, Q)$ pair and training target $A$ are observed. 
Such an absence motivates VideoQA to embrace video grounding in its modeling. 
Specifically, in light of question $Q$, the estimated causal scene $\hat{C}$ is grounded from the massive $V$ to approach the oracle $C$ and then generate prediction $\hat{A}$ via $Q \rightarrow A \leftarrow C$. 
To systematize this relation, the causal-equivariance principle introduces an equivariant transformation $T_\epsilon$ to each of the parent variables (\ie $C$ and $Q$), and expects a proportionate change in the response variable (\ie $A$). On top of SCM, we formally present such notions as:
\begin{gather}
    T_\epsilon(\hat{A})=f_{\hat{A}}(T_\epsilon(\hat{C}),T_\epsilon(Q)). \label{equ:equivariance}
\end{gather}
Meanwhile, environment-invariant principle formulated Equation \eqref{equ:eq0} in the sense that imposing an invariant-transformation $T_\iota$ on the estimated environment $\hat{E}$ should not trigger variation of answer $A$: 
\begin{gather}
      \hat{A}=f_{\hat{A}}(T_\iota(\hat{E}),Q)),\label{equ:invariance}
\end{gather} 
To this end, we parameterize our learning framework, EIGV, as a combination of equivariant and invariant principles, which comprises three additional modules on top of the ERM-guided backbone: grounding indicator, intervener, and disruptor. In a nutshell, we display our EIGV framework in Figure \ref{fig:main}.

\vspace{5pt}
\noindent \textbf{Data representation.}
Following previous efforts \cite{jiang2020reasoning, DBLP:conf/acl/GuoZJ0L20}, we encode video instance $v$ as a sequence of $K$ fixed visual clips, while question instance $q$ is encoded into a similar form with a fixed length of language tokens $L$.
Then, visual and linguistic features are applied with a linear layer and an LSTM \cite{10.1162/neco.1997.9.8.1735}, respectively, to align their dimension. As a result, we acquire the output of linear layer $ \vb{v} \in \mathbb{R}^{k \times d}$ as the final video representation and the last hidden state of LSTM $\vb{q} \in \mathbb{R}^{d}$ as the holistic question representation.


\subsection{Grounding Indicator}
Scene partition is fundamental to the rationale discovery, whose core is to estimate the value of $C$ and $E$ via a hard split on video $V$. Given an input sample $(v,q)$, the grounding indicator aims to access the causal scene and environment scene via their estimated value $\hat{c}$ and $\hat{e}$ according to question $Q$. 
Concretely, we first construct two cross-modal attention modules to indicate the probability of each visual clip of being causal scene ($\vb{p}_{\hat{c}} \in\mathbb{R}^{K}$) 
and environment scene ($\vb{p}_{\hat{e}}\in\mathbb{R}^{K}$):
\begin{align}
    \vb{p}_{\hat{c}} &= \text{Softmax}(\text{FC}_{1}(\vb{v})\cdot\text{FC}_{2}(\vb{q})^\intercal),\\
    \vb{p}_{\hat{t}} &= \text{Softmax}(\text{FC}_{3}(\vb{v})\cdot\text{FC}_{4}(\vb{q})^\intercal),
\end{align}
where $\text{FC}_{1},\text{FC}_{2},\text{FC}_{3},\text{FC}_{4}$ are fully connected layers that align cross-modal representations.
However, gathering messages via a soft mask still makes the visual information on different clips overlap. 
As discussed in Section 3.2, guided by ERM, the conventional attention mechanism is unable to block the influence of $\hat{e}$, thus undermining the veracity of $\hat{c}$.
%
As a correction, the grounding indicator makes a discrete selection over the clip-wise attention result to generate a disjoint group of the causal scene. We leverage Gumbel-Softmax \cite{DBLP:conf/iclr/JangGP17} to manage a differentiable selection on attentive probabilities and compute the indicator vector $\vb{I}\in\mathbb{R}^{K\times 2}$ on the two attention scores  over each clip (\ie $\vb{p}_{\hat{c,i}}$, $\vb{p}_{\hat{e,i}}$, $i \in K$).  Formally, $\vb{I}$ is derived as:
\begin{gather}
    \vb{I} = \text{Gumbel-Softmax}([\vb{p}_{\hat{c}};\vb{p}_{\hat{e}}]), 
\end{gather}
where $[;]$ denote concatenation. The first and second column of $\vb{I}$ (\ie $I_{0}$ and $I_{1}$) index the attribution of $\hat{c}$ and $\hat{e}$ over k clips, respectively. 
To this end, we estimate $\hat{c}$ and $\hat{e}$ as follows:
\begin{gather}
    \hat{c} = I_{0}\cdot v ,\quad \hat{e} = I_{1}\cdot v, \quad \st v=\hat{c}+\hat{e} .
\end{gather}


\subsection{Intervener}
\begin{figure}[t]
\centering
\includegraphics[scale=0.43]{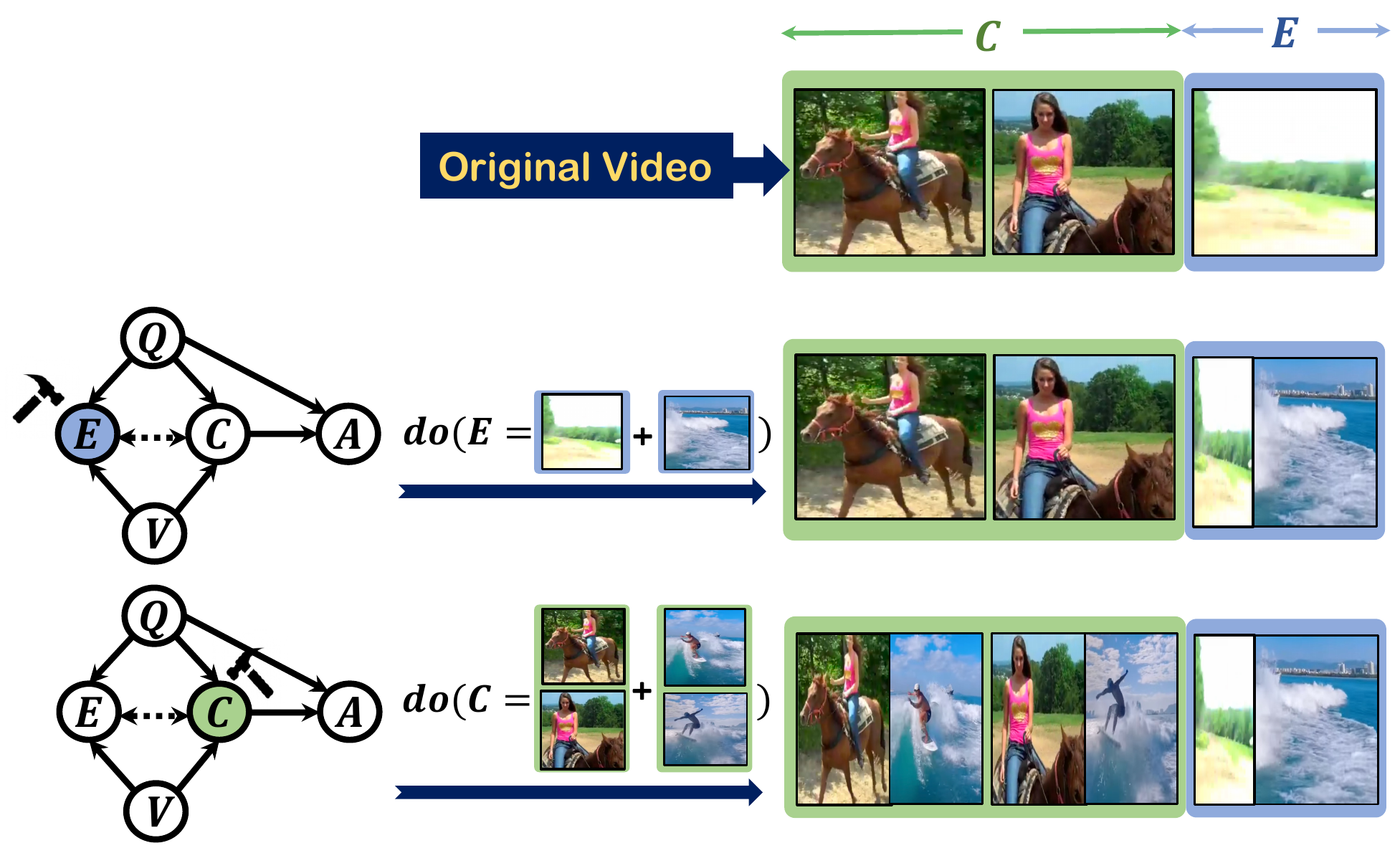}
\caption{We illustrate the invariant and equivariant interventions in the second and third rows, respectively. The effects on $Q$ and $A$ are omitted for illustration purposes.}
\vspace{-5pt}
\label{fig:intervene}
\end{figure}

In absence of clip-level supervision, learning grounding indicators requires dedicated exploitation of the equivariance-invariance principle.
On this demand, we propose the intervener, which prompts the estimated rationale to the oracle by intervening $\hat{c}$ and $\hat{e}$. 
Figure \ref{fig:intervene} describes the functionality of $do(\cdot)$ --- the intervention operator that successively manipulated SCM over $E$ and $C$. Concretely, two intervention operations are configured to realize the equivariant and invariant transformation defined in Equations \eqref{equ:equivariance} and \eqref{equ:invariance}. 
%

To fulfill the causal-equivariant principle, we design the E-intervention on the causal scene $\hat{c}$, which applies a linear interpolation between two data points on their causal factors --- $C$, $Q$ and $A$.  
By casting the same mixing ratio $\lambda_0\sim\text{Beta}(\alpha,\alpha)$ on all causal factors, the equivariant intervener learns to capture 
the causal connection of $C, Q \to A$.
In particular, we attain the intervened causal scene $c^*\in \mathbb{R}^{K\times d}$, question $q^*\in \mathbb{R}^{d}$ and answer $a^*\in \mathbb{R}$ as follow:
\begin{gather}
    c^*=\lambda_0 \cdot \hat{c}+(1-\lambda_0) \cdot \hat{c}',\\
    q^*=\lambda_0 \cdot q+(1-\lambda_0) \cdot q',\\
    a^*=\lambda_0 \cdot a+(1-\lambda_0) \cdot a',
\end{gather}
where $\hat{c}'$, $q'$ and $a'$ are causal factors from a second sample.

To achieve the environment-invariant principle, we devise the I-intervention that adopts a similar mixing strategy to the environment scene $\hat{e}$. 
Notably, by drawing the mixing ratios $\lambda_1$ from a second distribution that is distinct from the equivariant one (\ie $\lambda_1\sim\text{U}(0,1)$), the invariant intervention learns to rule out the influence of environment scene on the answer, which essentially refines the ERM-guided scheme at our will. 
Formally, we arrive the intervened environment scene $e^*$ by:
\begin{gather}
    e^*=\lambda_1 \cdot \hat{e}+(1-\lambda_1) \cdot \hat{e}',
\end{gather}
where $\hat{e}'$ is the estimated environment scene of a second sample.

In practice, the equvariant and invariant intervention operations are performed in parallel on different parts of $v$, and the intervened video $v^* \in \mathbb{R}^{K\times d}$ is composed of $do(C=c^*)$ and $do(E=e^*)$:
\begin{gather}
    v^*=c^*+e^*.
\end{gather}


%
\subsection{Disruptor}

To fully exploit the privilege of the proposed principles, we employ contrastive learning as an auxiliary objective to establish a good representation that maintains the desired properties of $\hat{c}$ and $\hat{e}$. 
Specifically, we first compose a memory bank $\pi$ as a collection of visual clips from other training videos. Then, we apply the grounding indicator a second time on top of the intervened variables, where the re-grounded causal and environment scene are manipulated to set up the contrastive twins as follows:
\begin{itemize}[leftmargin=*]
\item In light of the environment-invariance principle, positive video is developed in the sense that changing the environment scene will not provoke disagreement in answer semantics.  Thus, the disruptor synthesizes a positive video $v^+$ by disrupting the $v^*$ on its environment part --- that is, replacing the environment scene with a random stratification sampled from the memory bank.\footnote{Note that the environment substitutes will not involve the question-relevant scenes, to avoid creating additional paths from $E$ to $A$.}  
\item Built upon the causal-equivariance principle, the negative counterpart $v^-$ is created by a similar disruption but on the causal scene of $v^*$, where substitution on the question-critical causal part should raise inconsistency in answer space.
%
Apart from the visual negatives, the disruptor also creates linguistic alternatives to enhance the distinctiveness of the vision-language alignment. Specifically, it disrupts the combination of the intervened input ($v^*, q^*$) and pairs the video with random sample question $q_r$ to create a second view of negative samples ($v^*, q_r$).
\end{itemize}
To this end, we attain the answer representation of anchor $\vb{a}$ and its contrastive counterparts $\vb{a}^+, \vb{a}^-$ by feeding the paired positive and negative samples to backbone VideoQA model $f_{\hat{A}}$: 
\begin{gather}
    \vb{a}=f_{\hat{A}}(v^*,q^*),\\
    \vb{a}^+=f_{\hat{A}}(v^+,q^*),\\
    \vb{a}^-=f_{\hat{A}}([(v^-,q^*);\,(v^*,q_r)]),
\end{gather}
where $[;]$ denotes concatenation.

Notably, EIGV is designed to be model-agnostic, which aims to promote any VideoQA backbone built on frame-level visual inputs.
%

\subsection{Optimization}
By far, we set up the intervened vision-language instance ($v^*,q^*$) for a pair of input ($v,q$), and further constitute its contrastive counterparts based on the estimated grounding result. 
To steer the learning process away from the conventional ERM pitfall, we establish two learning objectives on top of their output $a, a^+, a^-$:

\begin{itemize}[leftmargin=*]
    \vspace{5pt}
    \item \noindent \textbf{Contrastive loss.} To reflect the invariant property of the environment scene while maintaining the distinctiveness of the causal scene, we borrow the definition of InfoNCE \cite{DBLP:journals/corr/abs-1807-03748}, and construct the contrastive objective as follows:
    \begin{gather}
        \mathcal{L}_{CL}=-log(\frac{\text{exp}{(a^\top a^+)}}{\text{exp}{(a^\top a^+)}\,+\sum_{n}^{N}\text{exp}{(a^\top a_n^-)}}),
    \end{gather}
    where $N$ is the number of negative samples, \lyc{$a_n^-$ denotes negative anwer generated by one of  negative samples.} 
    
    \vspace{5pt}
    \item \noindent \textbf{ERM loss.} 
    Estimating the rationale requires a robust causal connection from $V,Q$ to $A$. Thus, we imposed an entropy-based risk function \lyc{$\text{XE}(\cdot)$} on $(v^*,q^*)$ to approach the intervened answer $a^*$:
    \begin{gather}
        \mathcal{L}_{ERM}=\text{XE}(f_{\hat{A}}(v^*,q^*), a^*),
    \end{gather}
\end{itemize}

As a result, the overall training objective of EIGV is the aggregation of the forgoing risks:
\begin{gather}
   \mathcal{L}_{\text{EIGV}} = \mathbb{E}_{(v,q,a)\in\mathcal{O}}\mathcal{L}_{ERM} + \beta\mathcal{L}_{CL},
\end{gather}
where $\mathcal{O}$ is the set of training instances $(v,q)$ alongside their ground-truth answer $a$; $\beta$ is the hyper-parameter that balances the strength of contrastive learning. The joint optimization disentangles the mischief of environment scene, thus fulfilling the desired interpretation by locating the causal pattern. During inference, EIGV generates the predication $\hat{a}$ without the intervener and disruptor involved, and gives the interpretation $\hat{c}$ as the partition result of the grounding indicator.




\section{Experiment}
\label{sec:experiment}

In this section, we show the experimental results to answer the following research questions.
\begin{itemize}[leftmargin=*]
\item \textbf{RQ1} How effective is EIGV in discovering the causal pattern and improving the model generalization across different settings?
\item \textbf{RQ2} How does the sub-module and feature setting contribute to the performance?
\item \textbf{RQ3} What pattern does EIGV capture in rationale discovery?
\end{itemize}

\subsection{Settings}

\vspace{5pt}
\noindent \textbf{Datasets.} We conduct experiments on three benchmark datasets that challenge the model's reasoning capacity from different aspects: 
\textbf{MSVD-QA}  \cite{DBLP:conf/mm/XuZX0Z0Z17} and \textbf{MSRVTT-QA} \cite{DBLP:conf/mm/XuZX0Z0Z17} mainly emphasize the recognition ability by asking the descriptive questions, where 50K and 243K question-answer pairs are automatically generated from the human-labeled video captions, respectively.
\textbf{NExT-QA} \cite{DBLP:conf/cvpr/XiaoSYC21} pinpoints the causal and temporal relations among objects in the video. It contains 47.7K questions with answers in the form of multi-choice, which are manually annotated from 5.4K videos.

\vspace{5pt}
\noindent\textbf{Baseline.} We validate the effectiveness of EIGV across backbone VideoQA models of three kinds: 
1) \textbf{Memory-based} methods that foster a storage of input sequence via auxiliary memory design, such as AMU \cite{DBLP:conf/mm/XuZX0Z0Z17}, HME \cite{fan2019heterogeneous} and Co-Mem \cite{gao2018motionappearance}.
2) \textbf{Graph-based} methods that leverage the expressiveness of graph network to model the interaction between visual and language elements, which involves methods like L-GCN \cite{huang2020locationaware}, B2A  \cite{park2021bridge} and HGA \cite{jiang2020reasoning}. 
3) \textbf{Hierarchy-based} methods include HCRN \cite{le2021hierarchical}, PGAT \cite{DBLP:conf/mm/PengYBW21}, HOSTR \cite{dang2021hierarchical}, MSPAN \cite{DBLP:conf/acl/GuoZJ0L20} and HQGA \cite{xiao2021video}. In common, they exploit the multi-granularity nature of visual elements and realize the hierarchical reasoning via bottom-up architecture. 
In Specific, we test the generalization of EIGV by marrying our learning principles to three backbones of different categories: memory-based Co-Mem \cite{gao2018motionappearance}, graph-based HGA \cite{jiang2020reasoning} and hierarchy-based MSPAN \cite{DBLP:conf/acl/GuoZJ0L20}. 

\vspace{5pt}
\noindent \textbf{Implementation Detail.} 
For input representation, we encode the video instance as a sequence of $K$=16 clips, where each clip is represented as a combination of appearance and motion features extracted from the pre-trained ResNet-152 and 3D ResNeXt-101. For the linguistic feature, we follow \cite{DBLP:conf/cvpr/XiaoSYC21} and obtain the contextualized word representation using the fine-tuned BERT model. In the hyper-parameters setting, we set $d=512$ for cross-modal alignment, then train the model for 80 epochs with an initial learning rate of 5e-5.  During optimization, EIGV is trained with Adam optimizer and we decay the learning rate when validation stops improving for 5 epochs. The balance ratio $\beta$ is set to 0.75.

\subsection{Main Result (RQ1)}
\begin{table}[t]
  \centering
  \caption{Comparison with SoTAs. Our results are highlighted.}
    \resizebox{0.95\linewidth}{!}{
    \begin{tabular}{cc|ccc}
    \toprule
          & Model & MSVD-QA & MSRVTT-QA & NExT-QA \\
    \midrule
    \midrule
    \multirow{9}[6]{*}{\rot{SoTAs}} & AMU \cite{DBLP:conf/mm/XuZX0Z0Z17}   & 32.0    & 32.0    & - \\
          & HME \cite{fan2019heterogeneous} \  & 33.7  & 33.0    & 49.2 \\
\cmidrule{2-5}          & B2A  \cite{park2021bridge}   & 37.2  & 36.9  & - \\
          & L-GCN \cite{huang2020locationaware} & 34.3  & 33.7  & 49.5 \\
\cmidrule{2-5}          & HCRN \cite{le2021hierarchical}  & 36.1  & 35.6  & 48.9 \\
          & PGAT \cite{DBLP:conf/mm/PengYBW21} & 39.0    & 38.1  & - \\
          & HOSTR \cite{dang2021hierarchical} & 39.4  & 35.9  & - \\
          & \lyc{HQGA} \cite{xiao2021video} & \underline{41.2}  & \underline{38.6}  & \underline{51.8} \\
\cmidrule{2-5} & \lyc{IGV} \cite{Li_2022_CVPR} & 40.8  & 38.3  & 51.3 \\
    \midrule
    \midrule
    \multirow{6}[6]{*}{\rot{Ours}} & Co-Mem \cite{gao2018motionappearance}& 34.6  & 35.3  & 48.5 \\
          & \white{66666}$+$ EIGV   & \cellcolor[rgb]{ .851,  .851,  .851}         \textcolor{gray}{666}39.8 $^{\imp{+5.2}}$ & \cellcolor[rgb]{ .851,  .851,  .851} \textcolor{gray}{666}37.2 $^{\imp{+1.9}}$ & \cellcolor[rgb]{ .851,  .851,  .851} \textcolor{gray}{666}50.7 $^{\imp{+2.2}}$ \\
\cmidrule{2-5}          &  HGA \cite{jiang2020reasoning}   & 36.6  & 36.7  & 50.0 \\
          & \white{66666}$+$ EIGV   & \cellcolor[rgb]{ .851,  .851,  .851} \textcolor{gray}{666}40.8 $^{\imp{+4.2}}$& \cellcolor[rgb]{ .851,  .851,  .851} \textcolor{gray}{666}38.5 $^{\imp{+1.8}}$ & \cellcolor[rgb]{ .851,  .851,  .851} \textcolor{gray}{666}\textbf{53.7} $^{\imp{+3.7}}$ \\
\cmidrule{2-5}          & MSPAN \cite{DBLP:conf/acl/GuoZJ0L20} & 40.3  & 38.0    & 50.7 \\
          & \white{66666}$+$ EIGV   & \cellcolor[rgb]{ .851,  .851,  .851} \textcolor{gray}{666}\textbf{42.6} $^{\imp{+2.3}}$ & \cellcolor[rgb]{ .851,  .851,  .851} \textcolor{gray}{666}\textbf{39.3} $^{\imp{+1.3}}$ & \cellcolor[rgb]{ .851,  .851,  .851} \textcolor{gray}{666}52.9 $^{\imp{+2.2}}$ \\
    \bottomrule
    \end{tabular}%
    }
    \vspace{-5pt}
  \label{tab:main}%
\end{table}%

Table \ref{tab:main} shows the overall result of our method and the SoTAs on three benchmark datasets: MSVD-QA, MSRVTT-QA and NExT-QA. Our observations are summarized as follows:

\begin{itemize}[leftmargin=*]
\item Across all three benchmark datasets, the proposed EIGV outperforms SoTA by a distinct margin (+1.2\%$\sim$2.3\%). Such prevailing performance indicates the overall effectiveness of our design, which underpins the theoretical soundness of the equivariant and invariant principles. 

\item Narrowing the inspection to each of the three backbones, EIGV brings each backbone model a sharp gain across all benchmark datasets (+1.3\%$\sim$5.2\%), which evidences its model-agnostic property. 
Nevertheless, we notice that the improvements fluctuate across the backbones. As a comparison, on MSVD-QA and MSRVTT-QA benchmarks, EIGV acquires more favorable gains with backbone Co-Mem and HGA than it does with MSPAN. This is because the multi-granularity hierarchy empowers the MSPAN with robustness, especially to questions of the descriptive type. Therefore, it achieves stronger backbone performances on benchmarks that focus on the descriptive question (\ie MSVD-QA and MSRVTT-QA), which, in turn, account for the contribution of EIGV to some extent, thus makes improvement of MSPAN less remarkable.
In contrast, when it comes to the causal and temporal question (\ie NExT-QA) where the inherit advantage of MSPAN backbone vanishes,  EIGV shows equivalent improvements on all three backbones (+2.2\%$\sim$3.7\%). 
%

\item Comparing the average improvement across different benchmarks, we notice that EIGV achieves the best improvement on MSVD-QA (+2.3\%$\sim$5.2\%) while relatively moderate gains on MSRVTT-QA (+1.3\%$\sim$1.9\%) and NExT-QA (+2.2\%$\sim$3.7\%).
The reason for such discrepancy is that MSVD-QA is relatively small in size,
which constrains the reasoning ability of the backbone models by limiting their exposure to training instances.
As a comparison, MSVD-QA is five-time smaller than MSRVTT-QA in terms of QA pairs (43K vs 243K), and three-time smaller than NExT-QA in terms of video instances (1970 vs 5440).
However, such deficiency caters to the focal point of EIGV that develops better in a less generalized situation, thus leading to more preferable growth on MSVD-QA.
\end{itemize}

\begin{table}[t]
  \centering
  \caption{Evaluation on the effectiveness of sub-modules}
    \resizebox{0.9\linewidth}{!}{
    \begin{tabular}{c|cc|cc}
    \toprule
    \multirow{2}[2]{*}{Ablation} & \multicolumn{2}{c|}{MSVD-QA} & \multicolumn{2}{c}{NExT-QA} \\
          & MSPAN & HGA   & MSPAN & HGA \\
    \midrule
    \midrule
    SoTA Backbone & 40.3  & 36.6  & 50.7  & 50.0 \\
    \midrule
    $+$ Mixup \cite{DBLP:conf/iclr/ZhangCDL18} & 41.0    & 38.3  & 52.0    & 51.8 \\
    $+$ Intervener & 41.5  & 39.6  & 52.5  & 52.7 \\
    \midrule
    $+$ Disruptor & 40.9  & 37.6  & 51.0    & 51.1 \\
    \white{66666} $+$ Disrupt-Q & 40.6  & 37.0    & 50.8  & 51.0 \\
    \white{66666} $+$ Disrupt-V & 40.7  & 37.3  & 51.0    & 50.8 \\
    \midrule
    EIGV & \textbf{42.6} & \textbf{40.8} & \textbf{52.9} & \textbf{53.7} \\
    \bottomrule
    \end{tabular}%
    }
    \vspace{-15pt}
  \label{tab:ablation}%
\end{table}%

\subsection{In-Depth Study (RQ2)}
\vspace{5pt}
\subsubsection{\textbf{What are the effect of EIGV's components?}}

To comprehensively understand the reasoning mechanism of EIGV, we poke its structure with careful scrutiny. Specifically, we explore the effectiveness of the proposed intervener and disruptor by analyzing their performance with different backbones on two benchmarks. We report the corresponding performances in Table \ref{tab:ablation} and summarize our findings as follows:
\begin{itemize}[leftmargin=*]

\vspace{5pt}
\item \noindent \textbf{Effectiveness of Intervener.} \label{exp:intervener}
We first testify the substantial efficacy of the intervener by comparing its permanence ( ``$+$Intervener'' ) to the backbone. This brings constant gains across different settings (+1.2\%$\sim$3\%), which demonstrates the stability of our design. Then, we compare the result of the intervener with the conventional mixup augmentation \cite{DBLP:conf/iclr/ZhangCDL18}, which can be considered as a simplified case of the interventer that only applies the equivariant intervention to the entire training video. The result shows that our design outperforms the conventional mixup in all cases. This manifests that the benefit of invariant intervention is fundamental, and the functionality of invariance and equivariance principle are mutually reinforced.

\vspace{5pt}
\item \noindent \textbf{Effectiveness of Disruptor.} 
We validate the disruptor design by investigating its components --- the substitution on video (``$+$Disrupt-V'') and the permutation on question (``$+$Disrupt-Q''), respectively. 
Albeit moderate, improvement on (``$+$Disrupt-V'') shows that stressing causal scene can benefit visual robustness.
A similar trend also applies to ``$+$Disrupt-Q'' as well, the constant improvement in all settings shows that acknowledging artificial corrosion in ($v,q$) matching can strengthen vision-language alignment, which is in line with the current finding in the cross-model pre-train literature \cite{DBLP:conf/icml/RadfordKHRGASAM21}. 
Furthermore, the overall result on ``$+$Disrupt'' shows that the advancement of ``$+$Disrupt-V'' and ``$+$Disrupt-Q'' can be amplified by further integration. 
\end{itemize}

\begin{table}[t]
  \centering
  \caption{Study of feature setting. ``APP'' and ``MOT'' denotes using appearance and motion feature individually. }
    \resizebox{0.95\linewidth}{!}{
    \begin{tabular}{cc|cc|cc}
    \toprule
    \multicolumn{2}{c|}{\multirow{2}[2]{*}{Method}} & \multicolumn{2}{c|}{MSVD-QA} & \multicolumn{2}{c}{NExT-QA} \\
    \multicolumn{2}{c|}{} & MSPAN & HGA   & MSPAN & HGA \\
    \midrule
    \midrule
    \multirow{2}[2]{*}{\rot{APP}} & SoTA Backbone & 40.1  & 35.0    & 49.7  & 48.3 \\
          & \white{66666}$+$ EIGV   & \textbf{41.0}    & \textbf{39.5}  & \textbf{52.0}    & \textbf{52.4} \\
    \midrule
    \multirow{2}[2]{*}{\rot{MOT}} & SoTA Backbone & 37.8  & 33.6  & 49.4  & 47.5 \\
          & \white{66666}$+$ EIGV   & \textbf{39.3}  & \textbf{38.5}  & \textbf{51.1}  & \textbf{51.7} \\
    \bottomrule
    \end{tabular}%
    }
  \vspace{-15pt}
  \label{tab:unifeat}%
\end{table}%

\subsubsection{\textbf{What are the effects of different feature settings?}}
To answer this question, we perform uni-feature tests for the visual representation. Concretely, instead of combining the appearance and motion features together and then manipulating them as a whole, we run ablative experiments on each of them solely.  
As shown in Table \ref{tab:unifeat}, under all circumstances, EIGV can improve models trained with appearance and motion features in equivalent magnitude, even though the appearance feature is demonstrated to be more visually informative in backbone comparison. This verifies that our improvement is ascribed to both feature modes rather than accessing only one of them.


\subsubsection{\textbf{What are the effects of hyper-parameters?}}
\begin{figure}[t]
\centering
\includegraphics[scale=0.35]{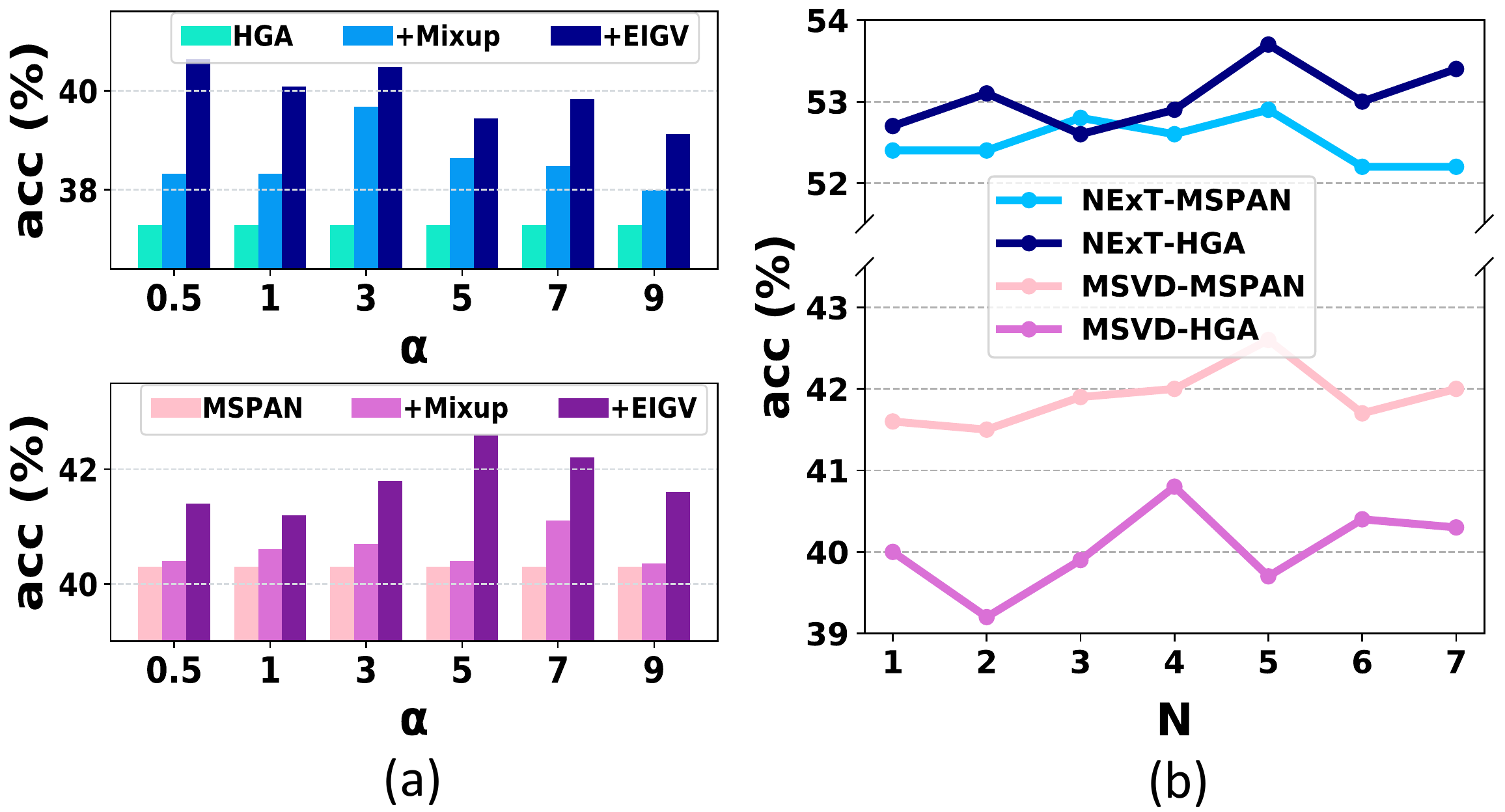}
\vspace{-10pt}
\caption{Hyperparameter analysis. (a) Study of $\alpha$ on MSVD-QA, which controls the equivariant mixing ratio by $\lambda_0\sim\text{Beta}(\alpha,\alpha)$. Performance of two EIGV enhanced models --- HGA (top) and MSPAN (bottom) are reported, alongside the SoTA backbone and mixup augmented performances.  
(b) Study on the impact of the negative sample number N, where EIGV with two backbones (\ie MSPAN and HGA) on two benchmark datasets (NExT-QA and MSVD-QA) are reported.}
\vspace{-10pt}
\label{fig:neg}
\end{figure}

Justifying a reliable design requires a sensitivity test on its hyper-parameters. As shown in Figure \ref{fig:neg}, we probe the potency of EIGV by investigating the distribution of the equivariance mixing ratio and the number of negative samples. Our observations are as follows:

\begin{itemize}[leftmargin=*]
\item Figure \ref{fig:neg}a shows how EIGV performs compared to the SoTA backbone and the conventional mixup augmentation. Specifically, we adjust $\alpha$ to vary the distribution that the equivariant mixing ratio $\lambda_0$ draws from, and cross-check the performance of EIGV (``+EIGV'') against its counterparts (``SoTA Backbone'' and ``+HGA'') on two backbone models. Mixup, despite some improvement, its generalization is limited by the choice of the backbone. For MSPAN backbone, even the heavily tuned $\alpha$ fails to make a reasonable improvement. In contrast, EIGV successively outperforms mixup augmentation and backbone methods in every $\alpha$ setting, which recalls our finding in Section \ref{exp:intervener} and justifies the necessity of the environment-invariance principle.      

\item Figure \ref{fig:neg}b demonstrates how the performance of EIGV varies as the number of negative sample increase. We notice that the predictive curves keep rising until $N$ reaches around five, which indicates that EIGV learns distinctive grounding rationale as more negative samples are considered. This is in line with the finding in the contrastive learning community that additional negative pairs bring about more desirable outcome \cite{he2019moco}. 
\end{itemize}

\subsection{Quantitative Study (RQ3)}
\begin{figure*}[t]
  \centering
  \includegraphics[width=0.97\textwidth]{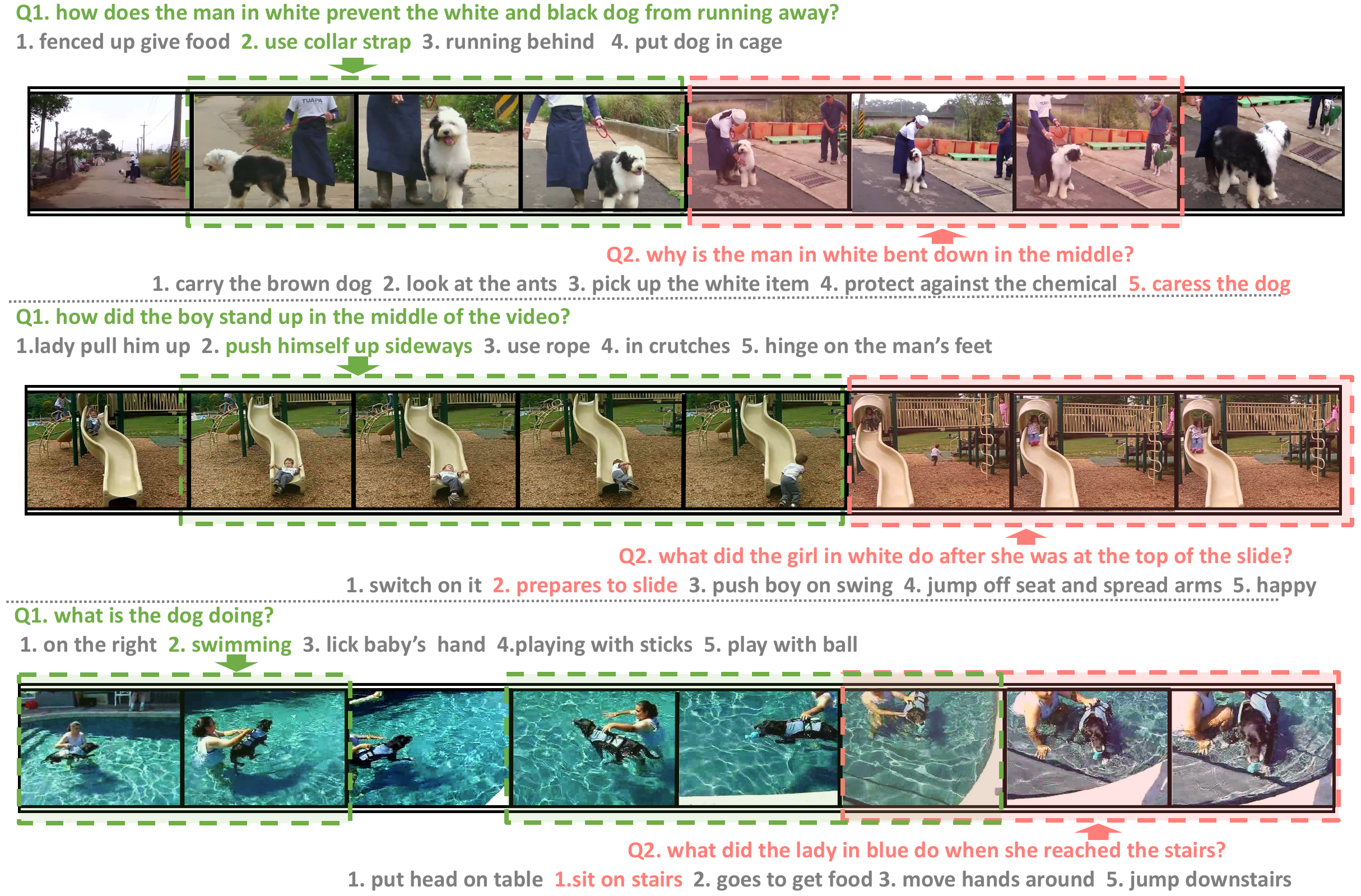}
 	\vspace{-13pt}
  \caption{Visualization of discovered grounding rationale. Each row comes with a video instance and two questions that target at different scene.  The \textcolor[rgb]{0,0.5,0}{green} and \textcolor[RGB]{225,152,150}{pink} windows indicate the rationales for the corresponding questions.}
  \label{fig:case-study}
 	\vspace{-10pt}
\end{figure*}

By nature, EIGV is equipped with intrinsic visual interpretability. To capture the learning insight of EIGV, we inspect the predictive answer of some video instances along with their grounded interpretations and show the visualization in Figure \ref{fig:case-study}, where each row provides a video instance and two questions that emphasize the visual content in different temporal spans. Notably, even for the same video instance, EIGV is able to accredit different scenes for the different questions. Nonetheless, we also observe the insufficient grounding result in Row 3 Q1, where the grounding result partially covers the dog swimming scene, while the whole video is answerable to the question.

\section{Related Work}
\label{sec:related}

\vspace{5pt}
\noindent\textbf{Video Question Answering.}
Established to answer the question in dynamic visual content, VideoQA is bred through the task of ImageQA but has broadened its definition by assembling a temporal dimension. 
To make the task intriguing, the VideoQA benchmark has gone beyond the problem of description \cite{DBLP:conf/mm/XuZX0Z0Z17} and built several datasets to challenge temporal reasoning and even causal reflection \cite{DBLP:conf/cvpr/XiaoSYC21}. 
As a result, VideoQA has experienced an aggressive expansion in the architecture design. 
Chronologically, early efforts tend to enact alignment through cross-modal attention \cite{zeng2016leveraging,li2019beyond} or enhanced memory design \cite{gao2018motionappearance, DBLP:conf/mm/XuZX0Z0Z17, fan2019heterogeneous}, while more recent works leverage the expressiveness of the graph neural network and perform the relation reasoning as node propagation \cite{jiang2020reasoning, DBLP:conf/mm/PengYBW21} or graph alignment \cite{park2021bridge}. 
In addition, current designs modify the representation of video and manipulate the temporal sequence from a hierarchical angle. Following this line, HCRN \cite{le2021hierarchical} first came out with the conditional relation module as building blocks that operate through different video intervals, whereas  HOSTR and PGAT made their advancement by incorporating visual content from different granularity. MSPAN, however, established cross-scale feature interaction on top of the hierarchy.
Despite effective, their intrinsic rationale has long been overlooked. To the best of our knowledge, EIGV is the first work that probes intrinsic interpretation. 

\vspace{5pt}
\noindent\textbf{Invariant Learning.}
Given a encoder $f(\cdot)$ and input $x$, a representation $f(x)$ is invariant to operation $T$, if $\forall x : f(G(x)) = f(x)$. 
In practice, this invariant property has a long history in presenting visual content (\eg HOG \cite{DBLP:conf/mmm/HuangTHTJ11}), which has recently been renovated by deep learning in form of risk minimization. As its most prevailing form, IRM \cite{arjovsky2020invariant} fosters this philosophy by posing an environment invariant prior and discovering the underlying causal pattern by reducing cross-environment variance.   
Different from previous studies that create environment via inductive re-grouping \cite{DBLP:conf/cvpr/AndersonWTB0S0G18} or adversarial inference \cite{DBLP:conf/icml/CreagerJZ21,wang2021causal, wang2022causal,wang2021clicks}, our method conducts causal intervention that perturbs the original sample distribution to form a new one.
\lyc{The most relvant work is \cite{Li_2022_CVPR}, where an invariant framework is introduced as a model-agnostic framework. However, EIGV gains better generalization ability by marrying equivariance as complementary learning principle.}

\vspace{5pt}
\noindent\textbf{Visual Interpretability}
Machine interpretability can be achieved in various methods, such as clustering \cite{monnier2020dticlustering} or disentanglement \cite{shen2020interfacegan}. Our design can be vested in the category of attribution discovery, which investigates the contribution of different input elements toward the prediction. 
Based on whether the prediction and interpretation are yielded simultaneously, two categories are further defined: 1) post-hoc methods that generated the interpretation after prediction, such as backpropagation methods (\eg grad-CAM \cite{DBLP:conf/iccv/SelvarajuCDVPB17}). 2) self-interpretable method that cast prediction and interpretation at the same stage.
Unlike the post-hoc method that traces the interpretative clue from the output of the black-box, the self-interpretable model builds a transparency model via methods such as prototype generation \cite{DBLP:journals/corr/abs-1806-10574} or structural delineation \cite{wu2022dir}. 
In fact, previous works tend to focus on static image. EIGV, however, approaches the video interpretation in a multi-modal situation. 

\vspace{-5pt}
\section{conclusion}
\label{sec:conclusion}
In this paper, we present EIGV --- a model-agnostic explainer, that empowers the SoTA VideoQA model with intrinsic interpretability. In the light of the causality, we formulate our learning principles --- causal-equivariance and environment-invariance by incorporating three constituents, the grounding indicator, the intervener, and the disruptor, which manage a robust rationale discovery. Experiments across three benchmarks validate EIGV's fulfillment in both interpretation and accuracy.



\bibliographystyle{ACM-Reference-Format}
\balance
\bibliography{sample-base}


\begin{thebibliography}{54}


\ifx \showCODEN    \undefined \def \showCODEN     #1{\unskip}     \fi
\ifx \showDOI      \undefined \def \showDOI       #1{#1}\fi
\ifx \showISBNx    \undefined \def \showISBNx     #1{\unskip}     \fi
\ifx \showISBNxiii \undefined \def \showISBNxiii  #1{\unskip}     \fi
\ifx \showISSN     \undefined \def \showISSN      #1{\unskip}     \fi
\ifx \showLCCN     \undefined \def \showLCCN      #1{\unskip}     \fi
\ifx \shownote     \undefined \def \shownote      #1{#1}          \fi
\ifx \showarticletitle \undefined \def \showarticletitle #1{#1}   \fi
\ifx \showURL      \undefined \def \showURL       {\relax}        \fi
\providecommand\bibfield[2]{#2}
\providecommand\bibinfo[2]{#2}
\providecommand\natexlab[1]{#1}
\providecommand\showeprint[2][]{arXiv:#2}

\bibitem[Anderson et~al\mbox{.}(2018)]%
        {DBLP:conf/cvpr/AndersonWTB0S0G18}
\bibfield{author}{\bibinfo{person}{Peter Anderson}, \bibinfo{person}{Qi Wu},
  \bibinfo{person}{Damien Teney}, \bibinfo{person}{Jake Bruce},
  \bibinfo{person}{Mark Johnson}, \bibinfo{person}{Niko S{\"{u}}nderhauf},
  \bibinfo{person}{Ian~D. Reid}, \bibinfo{person}{Stephen Gould}, {and}
  \bibinfo{person}{Anton van~den Hengel}.} \bibinfo{year}{2018}\natexlab{}.
\newblock \showarticletitle{Vision-and-Language Navigation: Interpreting
  Visually-Grounded Navigation Instructions in Real Environments}. In
  \bibinfo{booktitle}{\emph{CVPR}}. \bibinfo{pages}{3674--3683}.
\newblock


\bibitem[Arjovsky et~al\mbox{.}(2019)]%
        {arjovsky2020invariant}
\bibfield{author}{\bibinfo{person}{Mart{\'{\i}}n Arjovsky},
  \bibinfo{person}{L{\'{e}}on Bottou}, \bibinfo{person}{Ishaan Gulrajani},
  {and} \bibinfo{person}{David Lopez{-}Paz}.} \bibinfo{year}{2019}\natexlab{}.
\newblock \showarticletitle{Invariant Risk Minimization}.
\newblock \bibinfo{journal}{\emph{CoRR}}  \bibinfo{volume}{abs/1907.02893}
  (\bibinfo{year}{2019}).
\newblock


\bibitem[Chen et~al\mbox{.}(2018)]%
        {DBLP:journals/corr/abs-1806-10574}
\bibfield{author}{\bibinfo{person}{Chaofan Chen}, \bibinfo{person}{Oscar Li},
  \bibinfo{person}{Alina Barnett}, \bibinfo{person}{Jonathan Su}, {and}
  \bibinfo{person}{Cynthia Rudin}.} \bibinfo{year}{2018}\natexlab{}.
\newblock \showarticletitle{This looks like that: deep learning for
  interpretable image recognition}.
\newblock \bibinfo{journal}{\emph{CoRR}} (\bibinfo{year}{2018}),
  \bibinfo{pages}{8928--8939}.
\newblock


\bibitem[Chen et~al\mbox{.}(2020)]%
        {CSS}
\bibfield{author}{\bibinfo{person}{Long Chen}, \bibinfo{person}{Xin Yan},
  \bibinfo{person}{Jun Xiao}, \bibinfo{person}{Hanwang Zhang},
  \bibinfo{person}{Shiliang Pu}, {and} \bibinfo{person}{Yueting Zhuang}.}
  \bibinfo{year}{2020}\natexlab{}.
\newblock \showarticletitle{Counterfactual Samples Synthesizing for Robust
  Visual Question Answering}. In \bibinfo{booktitle}{\emph{CVPR}}.
  \bibinfo{pages}{10797--10806}.
\newblock


\bibitem[Creager et~al\mbox{.}(2021)]%
        {DBLP:conf/icml/CreagerJZ21}
\bibfield{author}{\bibinfo{person}{Elliot Creager},
  \bibinfo{person}{J{\"{o}}rn{-}Henrik Jacobsen}, {and}
  \bibinfo{person}{Richard~S. Zemel}.} \bibinfo{year}{2021}\natexlab{}.
\newblock \showarticletitle{Environment Inference for Invariant Learning}. In
  \bibinfo{booktitle}{\emph{ICML}}. \bibinfo{pages}{2189--2200}.
\newblock


\bibitem[Dang et~al\mbox{.}(2021)]%
        {dang2021hierarchical}
\bibfield{author}{\bibinfo{person}{Long~Hoang Dang}, \bibinfo{person}{Thao~Minh
  Le}, \bibinfo{person}{Vuong Le}, {and} \bibinfo{person}{Truyen Tran}.}
  \bibinfo{year}{2021}\natexlab{}.
\newblock \showarticletitle{Hierarchical Object-oriented Spatio-Temporal
  Reasoning for Video Question Answering}. In
  \bibinfo{booktitle}{\emph{IJCAI}}. \bibinfo{pages}{636--642}.
\newblock


\bibitem[Fan et~al\mbox{.}(2019)]%
        {fan2019heterogeneous}
\bibfield{author}{\bibinfo{person}{Chenyou Fan}, \bibinfo{person}{Xiaofan
  Zhang}, \bibinfo{person}{Shu Zhang}, \bibinfo{person}{Wensheng Wang},
  \bibinfo{person}{Chi Zhang}, {and} \bibinfo{person}{Heng Huang}.}
  \bibinfo{year}{2019}\natexlab{}.
\newblock \showarticletitle{Heterogeneous Memory Enhanced Multimodal Attention
  Model for Video Question Answering}. In \bibinfo{booktitle}{\emph{CVPR}}.
  \bibinfo{pages}{1999--2007}.
\newblock


\bibitem[Gao et~al\mbox{.}(2018)]%
        {gao2018motionappearance}
\bibfield{author}{\bibinfo{person}{Jiyang Gao}, \bibinfo{person}{Runzhou Ge},
  \bibinfo{person}{Kan Chen}, {and} \bibinfo{person}{Ram Nevatia}.}
  \bibinfo{year}{2018}\natexlab{}.
\newblock \showarticletitle{Motion-Appearance Co-Memory Networks for Video
  Question Answering}. In \bibinfo{booktitle}{\emph{CVPR}}.
  \bibinfo{pages}{6576--6585}.
\newblock


\bibitem[Ghorbani et~al\mbox{.}(2019)]%
        {ghorbani2019interpretation}
\bibfield{author}{\bibinfo{person}{Amirata Ghorbani}, \bibinfo{person}{Abubakar
  Abid}, {and} \bibinfo{person}{James Zou}.} \bibinfo{year}{2019}\natexlab{}.
\newblock \showarticletitle{Interpretation of neural networks is fragile}. In
  \bibinfo{booktitle}{\emph{AAAI}}. \bibinfo{pages}{3681--3688}.
\newblock


\bibitem[Guo et~al\mbox{.}(2021)]%
        {DBLP:conf/acl/GuoZJ0L20}
\bibfield{author}{\bibinfo{person}{Zhicheng Guo}, \bibinfo{person}{Jiaxuan
  Zhao}, \bibinfo{person}{Licheng Jiao}, \bibinfo{person}{Xu Liu}, {and}
  \bibinfo{person}{Lingling Li}.} \bibinfo{year}{2021}\natexlab{}.
\newblock \showarticletitle{Multi-Scale Progressive Attention Network for Video
  Question Answering}. In \bibinfo{booktitle}{\emph{ACL}}.
  \bibinfo{pages}{973--978}.
\newblock


\bibitem[He et~al\mbox{.}(2020)]%
        {he2019moco}
\bibfield{author}{\bibinfo{person}{Kaiming He}, \bibinfo{person}{Haoqi Fan},
  \bibinfo{person}{Yuxin Wu}, \bibinfo{person}{Saining Xie}, {and}
  \bibinfo{person}{Ross~B. Girshick}.} \bibinfo{year}{2020}\natexlab{}.
\newblock \showarticletitle{Momentum Contrast for Unsupervised Visual
  Representation Learning}. In \bibinfo{booktitle}{\emph{CVPR}}.
  \bibinfo{pages}{9726--9735}.
\newblock


\bibitem[Heo et~al\mbox{.}(2019)]%
        {heo2019fooling}
\bibfield{author}{\bibinfo{person}{Juyeon Heo}, \bibinfo{person}{Sunghwan Joo},
  {and} \bibinfo{person}{Taesup Moon}.} \bibinfo{year}{2019}\natexlab{}.
\newblock \showarticletitle{Fooling Neural Network Interpretations via
  Adversarial Model Manipulation}. In \bibinfo{booktitle}{\emph{NeurIPS}}.
  \bibinfo{pages}{2921--2932}.
\newblock


\bibitem[Hochreiter and Schmidhuber(1997)]%
        {10.1162/neco.1997.9.8.1735}
\bibfield{author}{\bibinfo{person}{Sepp Hochreiter} {and}
  \bibinfo{person}{Jürgen Schmidhuber}.} \bibinfo{year}{1997}\natexlab{}.
\newblock \showarticletitle{{Long Short-Term Memory}}.
\newblock \bibinfo{journal}{\emph{Neural Computation}} (\bibinfo{year}{1997}),
  \bibinfo{pages}{1735--1780}.
\newblock


\bibitem[Huang et~al\mbox{.}(2020)]%
        {huang2020locationaware}
\bibfield{author}{\bibinfo{person}{Deng Huang}, \bibinfo{person}{Peihao Chen},
  \bibinfo{person}{Runhao Zeng}, \bibinfo{person}{Qing Du},
  \bibinfo{person}{Mingkui Tan}, {and} \bibinfo{person}{Chuang Gan}.}
  \bibinfo{year}{2020}\natexlab{}.
\newblock \showarticletitle{Location-Aware Graph Convolutional Networks for
  Video Question Answering}. In \bibinfo{booktitle}{\emph{AAAI}}.
  \bibinfo{pages}{11021--11028}.
\newblock


\bibitem[Huang et~al\mbox{.}(2011)]%
        {DBLP:conf/mmm/HuangTHTJ11}
\bibfield{author}{\bibinfo{person}{Shih{-}Shinh Huang},
  \bibinfo{person}{Hsin{-}Ming Tsai}, \bibinfo{person}{Pei{-}Yung Hsiao},
  \bibinfo{person}{Meng{-}Qui Tu}, {and} \bibinfo{person}{Er{-}Liang Jian}.}
  \bibinfo{year}{2011}\natexlab{}.
\newblock \showarticletitle{Combining Histograms of Oriented Gradients with
  Global Feature for Human Detection}. In \bibinfo{booktitle}{\emph{MMM}}.
  \bibinfo{pages}{208--218}.
\newblock


\bibitem[Jang et~al\mbox{.}(2017)]%
        {DBLP:conf/iclr/JangGP17}
\bibfield{author}{\bibinfo{person}{Eric Jang}, \bibinfo{person}{Shixiang Gu},
  {and} \bibinfo{person}{Ben Poole}.} \bibinfo{year}{2017}\natexlab{}.
\newblock \showarticletitle{Categorical Reparameterization with
  Gumbel-Softmax}. In \bibinfo{booktitle}{\emph{ICLR}}.
\newblock


\bibitem[Jiang and Han(2020)]%
        {jiang2020reasoning}
\bibfield{author}{\bibinfo{person}{Pin Jiang} {and} \bibinfo{person}{Yahong
  Han}.} \bibinfo{year}{2020}\natexlab{}.
\newblock \showarticletitle{Reasoning with Heterogeneous Graph Alignment for
  Video Question Answering}. In \bibinfo{booktitle}{\emph{AAAI}}.
  \bibinfo{pages}{11109--11116}.
\newblock


\bibitem[Laugel et~al\mbox{.}(2019)]%
        {DBLP:conf/ijcai/LaugelLMRD19}
\bibfield{author}{\bibinfo{person}{Thibault Laugel},
  \bibinfo{person}{Marie{-}Jeanne Lesot}, \bibinfo{person}{Christophe Marsala},
  \bibinfo{person}{Xavier Renard}, {and} \bibinfo{person}{Marcin Detyniecki}.}
  \bibinfo{year}{2019}\natexlab{}.
\newblock \showarticletitle{The Dangers of Post-hoc Interpretability:
  Unjustified Counterfactual Explanations}. In
  \bibinfo{booktitle}{\emph{IJCAI}}. \bibinfo{pages}{2801--2807}.
\newblock


\bibitem[Le et~al\mbox{.}(2021)]%
        {le2021hierarchical}
\bibfield{author}{\bibinfo{person}{Thao~Minh Le}, \bibinfo{person}{Vuong Le},
  \bibinfo{person}{Svetha Venkatesh}, {and} \bibinfo{person}{Truyen Tran}.}
  \bibinfo{year}{2021}\natexlab{}.
\newblock \showarticletitle{Hierarchical Conditional Relation Networks for
  Multimodal Video Question Answering}.
\newblock \bibinfo{journal}{\emph{Int. J. Comput. Vis.}}  \bibinfo{volume}{129}
  (\bibinfo{year}{2021}), \bibinfo{pages}{3027--3050}.
\newblock


\bibitem[Li et~al\mbox{.}(2019)]%
        {li2019beyond}
\bibfield{author}{\bibinfo{person}{Xiangpeng Li}, \bibinfo{person}{Jingkuan
  Song}, \bibinfo{person}{Lianli Gao}, \bibinfo{person}{Xianglong Liu},
  \bibinfo{person}{Wenbing Huang}, \bibinfo{person}{Xiangnan He}, {and}
  \bibinfo{person}{Chuang Gan}.} \bibinfo{year}{2019}\natexlab{}.
\newblock \showarticletitle{Beyond RNNs: Positional Self-Attention with
  Co-Attention for Video Question Answering}. In
  \bibinfo{booktitle}{\emph{AAAI}}. \bibinfo{pages}{8658--8665}.
\newblock


\bibitem[Li et~al\mbox{.}(2022)]%
        {Li_2022_CVPR}
\bibfield{author}{\bibinfo{person}{Yicong Li}, \bibinfo{person}{Xiang Wang},
  \bibinfo{person}{Junbin Xiao}, \bibinfo{person}{Wei Ji}, {and}
  \bibinfo{person}{Tat-Seng Chua}.} \bibinfo{year}{2022}\natexlab{}.
\newblock \showarticletitle{Invariant Grounding for Video Question Answering}.
  In \bibinfo{booktitle}{\emph{CVPR}}. \bibinfo{pages}{2928--2937}.
\newblock


\bibitem[Li et~al\mbox{.}(2021)]%
        {li2021interventional}
\bibfield{author}{\bibinfo{person}{Yicong Li}, \bibinfo{person}{Xun Yang},
  \bibinfo{person}{Xindi Shang}, {and} \bibinfo{person}{Tat-Seng Chua}.}
  \bibinfo{year}{2021}\natexlab{}.
\newblock \showarticletitle{Interventional video relation detection}. In
  \bibinfo{booktitle}{\emph{ACM MM}}. \bibinfo{pages}{4091--4099}.
\newblock


\bibitem[Liu et~al\mbox{.}(2021)]%
        {DBLP:conf/iccv/Liu0WL21}
\bibfield{author}{\bibinfo{person}{Fei Liu}, \bibinfo{person}{Jing Liu},
  \bibinfo{person}{Weining Wang}, {and} \bibinfo{person}{Hanqing Lu}.}
  \bibinfo{year}{2021}\natexlab{}.
\newblock \showarticletitle{{HAIR:} Hierarchical Visual-Semantic Relational
  Reasoning for Video Question Answering}. In \bibinfo{booktitle}{\emph{ICCV}}.
  \bibinfo{pages}{1678--1687}.
\newblock


\bibitem[Monnier et~al\mbox{.}(2020)]%
        {monnier2020dticlustering}
\bibfield{author}{\bibinfo{person}{Tom Monnier}, \bibinfo{person}{Thibault
  Groueix}, {and} \bibinfo{person}{Mathieu Aubry}.}
  \bibinfo{year}{2020}\natexlab{}.
\newblock \showarticletitle{Deep Transformation-Invariant Clustering}. In
  \bibinfo{booktitle}{\emph{NeurIPS}}.
\newblock


\bibitem[Niu et~al\mbox{.}(2021)]%
        {niu2020counterfactual}
\bibfield{author}{\bibinfo{person}{Yulei Niu}, \bibinfo{person}{Kaihua Tang},
  \bibinfo{person}{Hanwang Zhang}, \bibinfo{person}{Zhiwu Lu},
  \bibinfo{person}{Xian-Sheng Hua}, {and} \bibinfo{person}{Ji-Rong Wen}.}
  \bibinfo{year}{2021}\natexlab{}.
\newblock \showarticletitle{Counterfactual VQA: A Cause-Effect Look at Language
  Bias}. In \bibinfo{booktitle}{\emph{CVPR}}. \bibinfo{pages}{12700--12710}.
\newblock


\bibitem[Park et~al\mbox{.}(2021)]%
        {park2021bridge}
\bibfield{author}{\bibinfo{person}{Jungin Park}, \bibinfo{person}{Jiyoung Lee},
  {and} \bibinfo{person}{Kwanghoon Sohn}.} \bibinfo{year}{2021}\natexlab{}.
\newblock \showarticletitle{Bridge To Answer: Structure-Aware Graph Interaction
  Network for Video Question Answering}. In \bibinfo{booktitle}{\emph{CVPR}}.
  \bibinfo{pages}{15526--15535}.
\newblock


\bibitem[Pearl(2009a)]%
        {pearl2009causal}
\bibfield{author}{\bibinfo{person}{Judea Pearl}.}
  \bibinfo{year}{2009}\natexlab{a}.
\newblock \showarticletitle{Causal inference in statistics: An overview}.
\newblock \bibinfo{journal}{\emph{Statistics surveys}} (\bibinfo{year}{2009}),
  \bibinfo{pages}{96--146}.
\newblock


\bibitem[Pearl(2009b)]%
        {reason:Pearl09a}
\bibfield{author}{\bibinfo{person}{Judea Pearl}.}
  \bibinfo{year}{2009}\natexlab{b}.
\newblock \bibinfo{booktitle}{\emph{Causality: Models, Reasoning and Inference}
  (\bibinfo{edition}{2nd} ed.)}.
\newblock \bibinfo{publisher}{Cambridge University Press}.
\newblock


\bibitem[Pearl et~al\mbox{.}(2016)]%
        {pearl2016causal}
\bibfield{author}{\bibinfo{person}{Judea Pearl}, \bibinfo{person}{Madelyn
  Glymour}, {and} \bibinfo{person}{Nicholas~P Jewell}.}
  \bibinfo{year}{2016}\natexlab{}.
\newblock \bibinfo{booktitle}{\emph{Causal inference in statistics: A primer}}.
\newblock


\bibitem[Peng et~al\mbox{.}(2021)]%
        {DBLP:conf/mm/PengYBW21}
\bibfield{author}{\bibinfo{person}{Liang Peng}, \bibinfo{person}{Shuangji
  Yang}, \bibinfo{person}{Yi Bin}, {and} \bibinfo{person}{Guoqing Wang}.}
  \bibinfo{year}{2021}\natexlab{}.
\newblock \showarticletitle{Progressive Graph Attention Network for Video
  Question Answering}. In \bibinfo{booktitle}{\emph{ACM MM}}.
  \bibinfo{pages}{2871--2879}.
\newblock


\bibitem[Radford et~al\mbox{.}(2021)]%
        {DBLP:conf/icml/RadfordKHRGASAM21}
\bibfield{author}{\bibinfo{person}{Alec Radford}, \bibinfo{person}{Jong~Wook
  Kim}, \bibinfo{person}{Chris Hallacy}, \bibinfo{person}{Aditya Ramesh},
  \bibinfo{person}{Gabriel Goh}, \bibinfo{person}{Sandhini Agarwal},
  \bibinfo{person}{Girish Sastry}, \bibinfo{person}{Amanda Askell},
  \bibinfo{person}{Pamela Mishkin}, \bibinfo{person}{Jack Clark},
  \bibinfo{person}{Gretchen Krueger}, {and} \bibinfo{person}{Ilya Sutskever}.}
  \bibinfo{year}{2021}\natexlab{}.
\newblock \showarticletitle{Learning Transferable Visual Models From Natural
  Language Supervision}. In \bibinfo{booktitle}{\emph{ICML}}.
  \bibinfo{pages}{8748--8763}.
\newblock


\bibitem[Ribeiro et~al\mbox{.}(2016)]%
        {LIME}
\bibfield{author}{\bibinfo{person}{Marco~T{\'{u}}lio Ribeiro},
  \bibinfo{person}{Sameer Singh}, {and} \bibinfo{person}{Carlos Guestrin}.}
  \bibinfo{year}{2016}\natexlab{}.
\newblock \showarticletitle{"Why Should {I} Trust You?": Explaining the
  Predictions of Any Classifier}. In \bibinfo{booktitle}{\emph{KDD}}.
  \bibinfo{pages}{1135--1144}.
\newblock


\bibitem[Ross et~al\mbox{.}(2017)]%
        {DBLP:conf/ijcai/RossHD17}
\bibfield{author}{\bibinfo{person}{Andrew~Slavin Ross},
  \bibinfo{person}{Michael~C. Hughes}, {and} \bibinfo{person}{Finale
  Doshi{-}Velez}.} \bibinfo{year}{2017}\natexlab{}.
\newblock \showarticletitle{Right for the Right Reasons: Training
  Differentiable Models by Constraining their Explanations}. In
  \bibinfo{booktitle}{\emph{IJCAI}}. \bibinfo{pages}{2662--2670}.
\newblock


\bibitem[Rudin(2019)]%
        {DBLP:journals/natmi/Rudin19}
\bibfield{author}{\bibinfo{person}{Cynthia Rudin}.}
  \bibinfo{year}{2019}\natexlab{}.
\newblock \showarticletitle{Stop explaining black box machine learning models
  for high stakes decisions and use interpretable models instead}.
\newblock \bibinfo{journal}{\emph{Nature Machine Intelligence}}
  (\bibinfo{year}{2019}), \bibinfo{pages}{206--215}.
\newblock


\bibitem[Selvaraju et~al\mbox{.}(2017)]%
        {DBLP:conf/iccv/SelvarajuCDVPB17}
\bibfield{author}{\bibinfo{person}{Ramprasaath~R. Selvaraju},
  \bibinfo{person}{Michael Cogswell}, \bibinfo{person}{Abhishek Das},
  \bibinfo{person}{Ramakrishna Vedantam}, \bibinfo{person}{Devi Parikh}, {and}
  \bibinfo{person}{Dhruv Batra}.} \bibinfo{year}{2017}\natexlab{}.
\newblock \showarticletitle{Grad-CAM: Visual Explanations from Deep Networks
  via Gradient-Based Localization}. In \bibinfo{booktitle}{\emph{ICCV}}.
  \bibinfo{pages}{618--626}.
\newblock


\bibitem[Seo et~al\mbox{.}(2021)]%
        {DBLP:conf/acl/SeoKPZ20}
\bibfield{author}{\bibinfo{person}{Ahjeong Seo}, \bibinfo{person}{Gi{-}Cheon
  Kang}, \bibinfo{person}{Joonhan Park}, {and} \bibinfo{person}{Byoung{-}Tak
  Zhang}.} \bibinfo{year}{2021}\natexlab{}.
\newblock \showarticletitle{Attend What You Need: Motion-Appearance Synergistic
  Networks for Video Question Answering}. In \bibinfo{booktitle}{\emph{ACL}}.
  \bibinfo{pages}{6167--6177}.
\newblock


\bibitem[Shen et~al\mbox{.}(2020)]%
        {shen2020interfacegan}
\bibfield{author}{\bibinfo{person}{Yujun Shen}, \bibinfo{person}{Ceyuan Yang},
  \bibinfo{person}{Xiaoou Tang}, {and} \bibinfo{person}{Bolei Zhou}.}
  \bibinfo{year}{2020}\natexlab{}.
\newblock \showarticletitle{InterFaceGAN: Interpreting the Disentangled Face
  Representation Learned by GANs}.
\newblock \bibinfo{journal}{\emph{TPAMI}} (\bibinfo{year}{2020}),
  \bibinfo{pages}{2004--2018}.
\newblock


\bibitem[Slack et~al\mbox{.}(2020)]%
        {slack2020fooling}
\bibfield{author}{\bibinfo{person}{Dylan Slack}, \bibinfo{person}{Sophie
  Hilgard}, \bibinfo{person}{Emily Jia}, \bibinfo{person}{Sameer Singh}, {and}
  \bibinfo{person}{Himabindu Lakkaraju}.} \bibinfo{year}{2020}\natexlab{}.
\newblock \showarticletitle{Fooling lime and shap: Adversarial attacks on post
  hoc explanation methods}. In \bibinfo{booktitle}{\emph{AIES}}.
  \bibinfo{pages}{180--186}.
\newblock


\bibitem[Torralba and Efros(2011)]%
        {DBLP:conf/cvpr/TorralbaE11}
\bibfield{author}{\bibinfo{person}{Antonio Torralba} {and}
  \bibinfo{person}{Alexei~A. Efros}.} \bibinfo{year}{2011}\natexlab{}.
\newblock \showarticletitle{Unbiased look at dataset bias}. In
  \bibinfo{booktitle}{\emph{CVPR}}. \bibinfo{publisher}{{IEEE} Computer
  Society}, \bibinfo{pages}{1521--1528}.
\newblock


\bibitem[van~den Oord et~al\mbox{.}(2018)]%
        {DBLP:journals/corr/abs-1807-03748}
\bibfield{author}{\bibinfo{person}{A{\"{a}}ron van~den Oord},
  \bibinfo{person}{Yazhe Li}, {and} \bibinfo{person}{Oriol Vinyals}.}
  \bibinfo{year}{2018}\natexlab{}.
\newblock \showarticletitle{Representation Learning with Contrastive Predictive
  Coding}.
\newblock \bibinfo{journal}{\emph{CoRR}}  \bibinfo{volume}{abs/1807.03748}
  (\bibinfo{year}{2018}).
\newblock


\bibitem[Wang et~al\mbox{.}(2021c)]%
        {DBLP:conf/mm/WangG0W21}
\bibfield{author}{\bibinfo{person}{Hui Wang}, \bibinfo{person}{Dan Guo},
  \bibinfo{person}{Xian{-}Sheng Hua}, {and} \bibinfo{person}{Meng Wang}.}
  \bibinfo{year}{2021}\natexlab{c}.
\newblock \showarticletitle{Pairwise {VLAD} Interaction Network for Video
  Question Answering}. In \bibinfo{booktitle}{\emph{ACM MM}}.
  \bibinfo{pages}{5119--5127}.
\newblock


\bibitem[Wang et~al\mbox{.}(2021a)]%
        {2021}
\bibfield{author}{\bibinfo{person}{Jianyu Wang}, \bibinfo{person}{Bing{-}Kun
  Bao}, {and} \bibinfo{person}{Changsheng Xu}.}
  \bibinfo{year}{2021}\natexlab{a}.
\newblock \showarticletitle{DualVGR: {A} Dual-Visual Graph Reasoning Unit for
  Video Question Answering}.
\newblock \bibinfo{journal}{\emph{CoRR}}  \bibinfo{volume}{abs/2107.04768}
  (\bibinfo{year}{2021}).
\newblock


\bibitem[Wang et~al\mbox{.}(2021d)]%
        {wang2021causal}
\bibfield{author}{\bibinfo{person}{Tan Wang}, \bibinfo{person}{Chang Zhou},
  \bibinfo{person}{Qianru Sun}, {and} \bibinfo{person}{Hanwang Zhang}.}
  \bibinfo{year}{2021}\natexlab{d}.
\newblock \showarticletitle{Causal Attention for Unbiased Visual Recognition}.
  In \bibinfo{booktitle}{\emph{ICCV}}. \bibinfo{pages}{3071--3080}.
\newblock


\bibitem[Wang et~al\mbox{.}(2021b)]%
        {wang2021clicks}
\bibfield{author}{\bibinfo{person}{Wenjie Wang}, \bibinfo{person}{Fuli Feng},
  \bibinfo{person}{Xiangnan He}, \bibinfo{person}{Hanwang Zhang}, {and}
  \bibinfo{person}{Tat-Seng Chua}.} \bibinfo{year}{2021}\natexlab{b}.
\newblock \showarticletitle{Clicks can be cheating: Counterfactual
  recommendation for mitigating clickbait issue}. In
  \bibinfo{booktitle}{\emph{SIGIR}}. \bibinfo{pages}{1288--1297}.
\newblock


\bibitem[Wang et~al\mbox{.}(2022)]%
        {wang2022causal}
\bibfield{author}{\bibinfo{person}{Wenjie Wang}, \bibinfo{person}{Xinyu Lin},
  \bibinfo{person}{Fuli Feng}, \bibinfo{person}{Xiangnan He},
  \bibinfo{person}{Min Lin}, {and} \bibinfo{person}{Tat-Seng Chua}.}
  \bibinfo{year}{2022}\natexlab{}.
\newblock \showarticletitle{Causal Representation Learning for
  Out-of-Distribution Recommendation}. In \bibinfo{booktitle}{\emph{WWW}}.
  \bibinfo{pages}{3562--3571}.
\newblock


\bibitem[Wang and Gupta(2018)]%
        {Wang_2018_ECCV}
\bibfield{author}{\bibinfo{person}{Xiaolong Wang} {and}
  \bibinfo{person}{Abhinav Gupta}.} \bibinfo{year}{2018}\natexlab{}.
\newblock \showarticletitle{Videos as Space-Time Region Graphs}. In
  \bibinfo{booktitle}{\emph{ECCV}}. \bibinfo{pages}{413--431}.
\newblock


\bibitem[Wu et~al\mbox{.}(2022)]%
        {wu2022dir}
\bibfield{author}{\bibinfo{person}{Ying-Xin Wu}, \bibinfo{person}{Xiang Wang},
  \bibinfo{person}{An Zhang}, \bibinfo{person}{Xiangnan He}, {and}
  \bibinfo{person}{Tat seng Chua}.} \bibinfo{year}{2022}\natexlab{}.
\newblock \showarticletitle{Discovering Invariant Rationales for Graph Neural
  Networks}. In \bibinfo{booktitle}{\emph{ICLR}}.
  \bibinfo{pages}{18446--18458}.
\newblock


\bibitem[Xiao et~al\mbox{.}(2021)]%
        {DBLP:conf/cvpr/XiaoSYC21}
\bibfield{author}{\bibinfo{person}{Junbin Xiao}, \bibinfo{person}{Xindi Shang},
  \bibinfo{person}{Angela Yao}, {and} \bibinfo{person}{Tat{-}Seng Chua}.}
  \bibinfo{year}{2021}\natexlab{}.
\newblock \showarticletitle{NExT-QA: Next Phase of Question-Answering to
  Explaining Temporal Actions}. In \bibinfo{booktitle}{\emph{CVPR}}.
  \bibinfo{pages}{9777--9786}.
\newblock


\bibitem[Xiao et~al\mbox{.}(2022)]%
        {xiao2021video}
\bibfield{author}{\bibinfo{person}{Junbin Xiao}, \bibinfo{person}{Angela Yao},
  \bibinfo{person}{Zhiyuan Liu}, \bibinfo{person}{Yicong Li},
  \bibinfo{person}{Wei Ji}, {and} \bibinfo{person}{Tat-Seng Chua}.}
  \bibinfo{year}{2022}\natexlab{}.
\newblock \showarticletitle{Video as Conditional Graph Hierarchy for
  Multi-Granular Question Answering}. In \bibinfo{booktitle}{\emph{AAAI}}.
  \bibinfo{pages}{2804--2812}.
\newblock


\bibitem[Xu et~al\mbox{.}(2017)]%
        {DBLP:conf/mm/XuZX0Z0Z17}
\bibfield{author}{\bibinfo{person}{Dejing Xu}, \bibinfo{person}{Zhou Zhao},
  \bibinfo{person}{Jun Xiao}, \bibinfo{person}{Fei Wu},
  \bibinfo{person}{Hanwang Zhang}, \bibinfo{person}{Xiangnan He}, {and}
  \bibinfo{person}{Yueting Zhuang}.} \bibinfo{year}{2017}\natexlab{}.
\newblock \showarticletitle{Video Question Answering via Gradually Refined
  Attention over Appearance and Motion}. In \bibinfo{booktitle}{\emph{ACM MM}}.
  \bibinfo{pages}{1645--1653}.
\newblock


\bibitem[Yang et~al\mbox{.}(2021)]%
        {DBLP:conf/cvpr/YangZQ021}
\bibfield{author}{\bibinfo{person}{Xu Yang}, \bibinfo{person}{Hanwang Zhang},
  \bibinfo{person}{Guojun Qi}, {and} \bibinfo{person}{Jianfei Cai}.}
  \bibinfo{year}{2021}\natexlab{}.
\newblock \showarticletitle{Causal Attention for Vision-Language Tasks}. In
  \bibinfo{booktitle}{\emph{CVPR}}. \bibinfo{pages}{9847--9857}.
\newblock


\bibitem[Zeng et~al\mbox{.}(2017)]%
        {zeng2016leveraging}
\bibfield{author}{\bibinfo{person}{Kuo{-}Hao Zeng},
  \bibinfo{person}{Tseng{-}Hung Chen}, \bibinfo{person}{Ching{-}Yao Chuang},
  \bibinfo{person}{Yuan{-}Hong Liao}, \bibinfo{person}{Juan~Carlos Niebles},
  {and} \bibinfo{person}{Min Sun}.} \bibinfo{year}{2017}\natexlab{}.
\newblock \showarticletitle{Leveraging Video Descriptions to Learn Video
  Question Answering}. In \bibinfo{booktitle}{\emph{AAAI}}.
  \bibinfo{pages}{4334--4340}.
\newblock


\bibitem[Zhang et~al\mbox{.}(2018)]%
        {DBLP:conf/iclr/ZhangCDL18}
\bibfield{author}{\bibinfo{person}{Hongyi Zhang}, \bibinfo{person}{Moustapha
  Ciss{\'{e}}}, \bibinfo{person}{Yann~N. Dauphin}, {and} \bibinfo{person}{David
  Lopez{-}Paz}.} \bibinfo{year}{2018}\natexlab{}.
\newblock \showarticletitle{mixup: Beyond Empirical Risk Minimization}. In
  \bibinfo{booktitle}{\emph{ICLR}}.
\newblock


\bibitem[Zhong et~al\mbox{.}(2022)]%
        {zhong2022video}
\bibfield{author}{\bibinfo{person}{Yaoyao Zhong}, \bibinfo{person}{Wei Ji},
  \bibinfo{person}{Junbin Xiao}, \bibinfo{person}{Yicong Li},
  \bibinfo{person}{Weihong Deng}, {and} \bibinfo{person}{Tat-Seng Chua}.}
  \bibinfo{year}{2022}\natexlab{}.
\newblock \showarticletitle{Video Question Answering: Datasets, Algorithms and
  Challenges}.
\newblock \bibinfo{journal}{\emph{arXiv preprint arXiv:2203.01225}}
  (\bibinfo{year}{2022}).
\newblock


\end{thebibliography}


\end{sloppypar}
\end{document}